\pdfoutput=1

\documentclass[11pt]{article}

\usepackage[preprint]{acl}

\usepackage{times}
\usepackage{latexsym}
\usepackage{tabularx}
\usepackage{array}           %
\usepackage{geometry}  
\usepackage{algorithm}
\usepackage{algpseudocode}
\usepackage{amsmath}
\usepackage{amsfonts}

\usepackage[T1]{fontenc}

\usepackage[utf8]{inputenc}

\usepackage{microtype}

\usepackage{graphicx}
\usepackage{subcaption}

\usepackage{booktabs} %

\usepackage{hyperref}
\usepackage{colortbl}  %

\usepackage{inconsolata}

\usepackage{graphicx}

\usepackage{amsmath}
\usepackage{amssymb}
\usepackage{mathtools}
\usepackage{amsthm}

\usepackage{float}

\usepackage[capitalize,noabbrev]{cleveref}

\newcolumntype{L}{>{\columncolor{gray!20}}l}

\theoremstyle{plain}

\theoremstyle{definition}

\theoremstyle{remark}

\usepackage{enumitem}

\usepackage{booktabs}  %
\usepackage{longtable} %
\usepackage{tcolorbox}

\usepackage{xspace}
\usepackage{ifthen}

\newboolean{showcomments}
\setboolean{showcomments}{true} %
\ifthenelse{\boolean{showcomments}}{
  \newcommand{\nbc}[3]{
    {\colorbox{#3}{\bfseries\sffamily\scriptsize\textcolor{white}{#1}}}%
    {\textcolor{#3}{\sf\small$\blacktriangleright$\textit{#2}$\blacktriangleleft$}}}
  \newcommand{\todo}[1]{\nbc{TODO}{#1}{violet}\xspace}
}{
  \newcommand{\nbc}[3]{}
  \newcommand{\todo}[1]{}
}

\newcommand{\tool}{LangProBe\xspace}

\definecolor{model_color}{HTML}{E69F00}             %
\definecolor{model_program_color}{HTML}{0072B2}       %
\definecolor{model_optimizer_color}{HTML}{009E73}     %
\definecolor{model_program_optimizer_color}{HTML}{CC79A7}  %

\title{LangProBe: a  Language Programs Benchmark}

\author{
 \textbf{Shangyin Tan\textsuperscript{1}},\hspace{1mm}
 \textbf{Lakshya A Agrawal\textsuperscript{1}},\hspace{1mm}
 \textbf{Arnav Singhvi\textsuperscript{2}},\hspace{1mm}
 \textbf{Liheng Lai\textsuperscript{1}},\hspace{1mm}
 \textbf{Michael J. Ryan\textsuperscript{2}},\hspace{1mm}
\\
 \textbf{Dan Klein\textsuperscript{1}},\hspace{1mm}
 \textbf{Omar Khattab\textsuperscript{3}},\hspace{1mm}
 \textbf{Koushik Sen\textsuperscript{1}},\hspace{1mm}
 \textbf{Matei Zaharia\textsuperscript{1}}
\\
\\
 \textsuperscript{1}University of California, Berkeley,
 \textsuperscript{2}Stanford University,
 \textsuperscript{3}Databricks
}

\begin{document}
\maketitle

\begin{abstract}
Composing language models (LMs) into multi-step \textit{language programs} and automatically optimizing their modular prompts is now a mainstream paradigm for building AI systems, but the tradeoffs in this space have only scarcely been studied before.
We introduce \tool, the first large-scale benchmark for evaluating the architectures and optimization strategies for language programs, with over 2000 combinations of tasks, architectures, optimizers, and choices of LMs. Using \tool, we are the first to study the impact of program architectures and optimizers (and their compositions together and with different models) on tradeoffs of quality and cost.
We find that optimized language programs offer strong cost--quality Pareto improvement over raw calls to models, but simultaneously demonstrate that human judgment (or empirical decisions) about which compositions to pursue is still necessary for best performance. We will open source the code and evaluation data for \tool.

\end{abstract}

\section{Introduction}
\label{sec:intro}

Language models are now routinely used to build modular natural-language software systems that process data or serve bespoke applications. Such \textit{language programs} make highly structured language model calls, invoke external tools, and compose all of these into sophisticated systems. In contrast to using language models as user-facing agents, these approaches compose better-scoped LM capabilities~\cite{compaiblogpost}, offer the models structure around accessing tools or information~\cite{DBLP:conf/nips/LewisPPPKGKLYR020,DBLP:conf/nips/KhattabPZ21, DBLP:journals/corr/abs-2203-05115}, and systematically scale planning and search at inference time~\cite{snell2024scalingllmtesttimecompute,saadfalcon2024archonarchitecturesearchframework}. %

To support such programming and permit portability across models and tasks, recent work has introduced declarative languages for expressing these systems and automating prompting (or fine-tuning) for their modules. For example, given a task-specific objective, DSPy \cite{DBLP:journals/corr/abs-2212-14024,khattab2024dspy} and TextGrad \cite{yuksekgonul2024textgradautomaticdifferentiationtext} are frameworks that study \textit{optimizing} these programs, e.g. by composing techniques like bootstrapping few-shot examples or refining free-form instructions, seeking to align the whole system to a distribution of inputs or a nuanced task.

\begin{figure}%
    \centering
    \includegraphics[width=\linewidth]{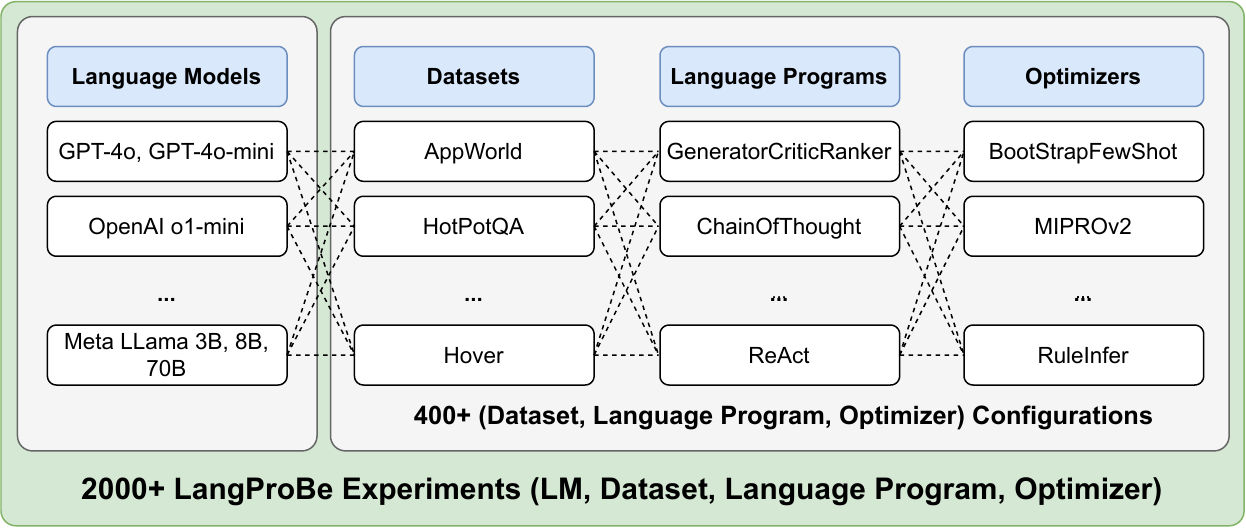}
    \caption{\tool includes 15 datasets, 4 optimizers, and more than 10 language programs, creating more than 400 configurations for evaluating language programs with different tasks and optimizers. }
    \label{fig:langprobe}
\end{figure}

Despite the interest in this space, it remains unclear which problems actually need modular programs, especially as models continue to improve, or which types of architectures and optimizers will work best for different problems. To begin to study this, we introduce \textbf{\tool} (\Cref{fig:langprobe}), a benchmark for evaluating combinations of language models, program architectures, and their optimizers. %
For many datasets, \tool implements multiple language programs (\Cref{subsec:bench_programs}), ranging from a single LM call to sophisticated and modular systems. Using \tool, we run over 2000 experiments to investigate several research questions.

\begin{figure*}[t!]
    \begin{subfigure}{0.98\linewidth}
        \includegraphics[width=\linewidth]{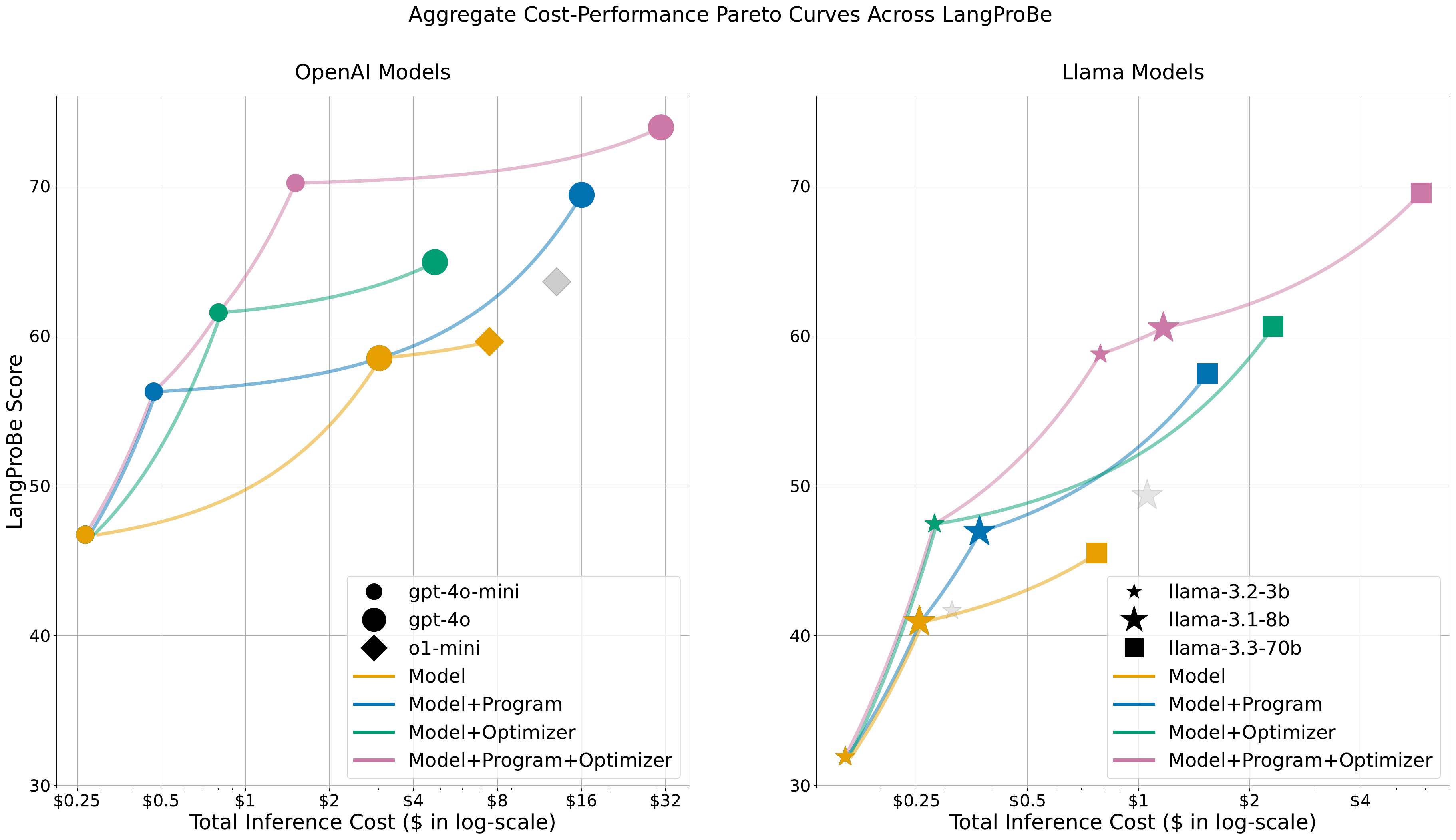}
    \end{subfigure}
    \caption{This figure shows stark cost-performance trade-offs across various configurations of (LM, Language Program, Optimizer), aggregated over multiple datasets in LangProBe. The Pareto curves represent the upper-left convex hull of achievable configurations. Piece-wise linear Pareto curve segments appear curved due to the log scale, but all points on the Pareto front are achievable via weighted (randomized) choice between the two endpoints.\\Four configurations are compared: 1) \textcolor{model_color}{Model}: Performance of baseline program (e.g., raw model predictions) without optimizers. 2) \textcolor{model_program_color}{Model+Program}: Performance with language programs applied, without optimizers. 3) \textcolor{model_optimizer_color}{Model+Optimizer}: Performance with optimizers applied to the baseline program. 4) \textcolor{model_program_optimizer_color}{Model+Program+Optimizer}: Performance of combined use of both language programs and optimizers. \textbf{Key Takeaway:} %
    For both model families, the \textcolor{model_program_optimizer_color}{Model+Program+Optimizer} Pareto curve deliver cost and quality improvements against \textcolor{model_program_color}{Model+Program} and \textcolor{model_optimizer_color}{Model+Optimizer} Pareto curves, which in turn improve over the \textcolor{model_color}{Model} Pareto curve, implying that using language programs and optimizing them can offer considerable gains not only with respect to quality, but also cost\protect\footnotemark. %
    }\label{fig:pareto} %
\end{figure*}

\begin{tcolorbox}
    Do language programs and optimizers for them improve cost-performance compared to using raw model calls? (\Cref{sec:eval_costs})
\end{tcolorbox}

On average, we find that language programs show non-trivial improvement over raw model prediction baselines even while considering language model costs. For example, the best optimized program running on gpt-4o-mini performs better than both the gpt-4o and o1-mini raw model prediction baselines at a significantly cheaper cost. However, this is far from uniform, as many self-contained problems (e.g., MMLU) can be straightforwardly tackled by powerful models without composition or optimization.

\footnotetext{Cost for OpenAI's models are from~\citep{openai_api_pricing} and for Meta's Llama models are from~\citep{AmazonBedrock2025}. Inference costs are likely to vary across time and providers, but provide a reasonable relative comparison between models by the same provider.}

\begin{tcolorbox}
    What language program architectures work best on different problems? (\Cref{sec:eval_programs})
\end{tcolorbox}

We find that modular programs are, perhaps expectedly, indispensable for tasks whose specification demands or strongly encourages access to external information or other tools. For example, tasks that require composing long-tail world knowledge benefit from retrieval-augmented generation (RAG) programs and multi-hop retrieval. While inference-time scaling programs are known to be helpful in certain cases, we find that they can fail to exceed baseline systems in certain applications, e.g. when error compounds across different modules. This illustrates a general theme we find, in which human judgment (or empirical decisions) about which compositions to pursue is still necessary for best performance.

\begin{tcolorbox}
    Which optimizers perform best and where?  (\Cref{sec:eval_optimizers})
\end{tcolorbox}

All optimizers can provide quality gains for many---though not all---combinations of models, tasks, and programs, but different optimizers create rich tradeoffs in terms of optimization and inference costs. On average, we find that MIPRO \cite{opsahlong2024optimizinginstructionsdemonstrationsmultistage}, an optimizer that constructs and explores combinations of instructions and few-shot examples through Bayesian search, performs best overall. As in prior work, we find that searching over combinations of \textit{self-bootstrapped} few-shot examples (BootstrapFewshotRandomSearch) is a highly competitive simple strategy, though the benefits of (bootstrapped) demonstrations heavily depend on the model--task pair. While this lies outside our scope, we believe that \tool defines a general methodology and large headroom for new prompt optimizers and also for finetuning- or RL-based optimizers to boost quality even further.

In summary, we contribute a \textbf{new benchmark for Language Programs} (\Cref{sec:langprobe}), the \textbf{first systematic investigation of the cost--quality tradeoffs of Language Programs} (\Cref{sec:eval_costs}), and new \textbf{empirical insights} showing that employing appropriate language programs with cheaper models for some tasks is both cost- and performance-optimal compared to using the raw prediction from a stronger model and offering guidance on which language program architecture and optimizer choices work best (\Cref{sec:eval_programs}, \Cref{sec:eval_optimizers}).

\section{Related Work}
Composing language model calls with tool usage into complex applications is now popular, with the help of established language model programming libraries such as DSPy \cite{DBLP:journals/corr/abs-2212-14024,khattab2024dspy}, LangChain \cite{Chase_LangChain_2022}, or TextGrad \cite{yuksekgonul2024textgradautomaticdifferentiationtext}. Language programs compose and integrate knowledge and information flow and are often equipped with additional reasoning capabilities. For instance, RAG \cite{DBLP:conf/nips/LewisPPPKGKLYR020} combines a language model with external retrieval from a corpus for knowledge-intensive tasks; Multi-Agent Debate \cite{du2023improvingfactualityreasoninglanguage} harnesses multiple models debating each other to sharpen mathematical and strategic reasoning; Self-Refine \cite{madaan2023selfrefineiterativerefinementselffeedback} iterates on its own outputs for continuous improvement; and ReAct \cite{yao2023reactsynergizingreasoningacting} integrates step-by-step reasoning with actions to facilitate interactions with external environments. Recent research also focuses on scaling inference-time computation by generating multiple responses and verifying them with specialized models \cite{snell2024scalingllmtesttimecompute, chen2024llmcallsneedscaling}.

Recent work has introduced agentic benchmarks \cite{kapoor2024aiagentsmatter, zhou2024webarenarealisticwebenvironment, yang2024sweagentagentcomputerinterfacesenable, wang2024novelqabenchmarkingquestionanswering}, which are closely related to \tool. As agents are a form of complex language programs, these benchmarks can help examine some design choices in language programs and we intend to add some of them to future iterations of \tool. Unfortunately, such agentic tasks leave plenty of unstudied problem types and are somewhat hard to adapt for asking questions about broader categories of program architectures~ \cite{stroebl2024inferencescalingflawslimits}. For example, SWE-bench \cite{yang2024sweagentagentcomputerinterfacesenable} is helpful in evaluating software engineer-like agents, but it can be difficult to rely on it for testing general inference-time scaling or retrieval-augmented generation methods. %

\begin{table*}[t!]
    \centering \small
    \renewcommand{\arraystretch}{1.2} %
    \setlength{\tabcolsep}{10pt}       %
    \begin{tabular}{L ll}              %
        \toprule
        \textbf{Dataset Category} & \textbf{Datasets \& Tasks} & \textbf{Specialized Programs} \\
        \midrule
        \textbf{Code}         & HumanEval, SWEUnderspecified,  & GeneratorCriticRanker, GeneratorCriticFuser\\
                              & SWEValidity                    & \\
        \midrule
        \textbf{Reasoning}    & Judge, Scone                   & GeneratorCriticRanker, GeneratorCriticFuser\\
        \midrule
        \textbf{Agent}        & AppWorld                     & ReActBaseline, ReActAugmented\\
        \midrule
        \textbf{Knowledge}    & MMLU, HoVer, IReRa, HotpotQA,   & RAGBasedRank, RAG, MultiHopSummarize,\\
                              & HotpotQAConditional, RAGQAArena & SimplifiedBaleen \\
        \midrule
        \textbf{Classification} & HeartDisease, Iris           & CoTBasedVote, GeneratorCriticRanker, \\ & & GeneratorCriticFuser\\
        \midrule
        \textbf{Math}         & MATH, GSM8K                  & GeneratorCriticRanker, GeneratorCriticFuser \\
        \bottomrule
    \end{tabular}
    \caption{Dataset categories, the datasets associated with them, and the specialized language programs evaluated on each different category. We also provide a detailed description for each dataset, task, and language program in \Cref{sec:appendix:bench} and \Cref{sec:appendix:language_program}.}
    \label{tab:datasets_by_category}
\end{table*}

\section{The LangProBe Benchmark} \label{sec:langprobe}
To offer an overview of \tool, we define different categories of benchmarks and then describe the selected programs and optimizers that we will study in this work. A detailed list of descriptions can be found in \Cref{sec:appendix:bench} and \Cref{sec:appendix:language_program}.

\subsection{Dataset Categories} \label{subsec:bench_category}
We choose datasets across diverse categories. In contrast to evaluations focused on comparing \textit{model capabilities}, e.g. HELM~\cite{liang2023holistic} or \texttt{lm-evaluation-harness}~\cite{eval-harness}, for each category, we recast existing datasets into a uniform testbed with metrics and data splits, seeking to establish a proxy for the bespoke applications that people program. Our categories include agentic tasks (AppWorld; \citealt{DBLP:conf/acl/TrivediKHMDLGSB24}), coding and software engineering tasks (SweBench annotation tasks; \citealt{OpenAISweAnnotation}, HumanEval; \citealt{chen2021evaluatinglargelanguagemodels}), mathematical and reasoning tasks (MATH; \citealt{DBLP:conf/nips/HendrycksBKABTS21}, GSM8K; \citealt{cobbe2021trainingverifierssolvemath}), domain-specific classification tasks (Iris; \citet{iris_53} and Heart Disease; \citet{heart_disease_45}), and question-answering problems (HotpotQA; \citealt{DBLP:conf/emnlp/Yang0ZBCSM18}, MMLU; \citealt{hendrycks2021measuringmassivemultitasklanguage}), and more.\footnote{We borrow a few programs and datasets like HotPotQA, HoVer, Iris, and Heart Disease from the open-source DSPy repository, contributed by \citet{khattab2024dspy} and \citet{opsahlong2024optimizinginstructionsdemonstrationsmultistage}, and adapt them to our more general setting.} %
While these serve an insightful starting point, we believe that future work must increasingly push \tool{} past the general capability datasets that model providers hill climb and closer to the composite downstream problems that LM programmers seek to solve.

\subsection{Language Programs} \label{subsec:bench_programs}

We adopt and design a diverse set of language program designs for the different tasks in \tool. By nature, language programs are much more structured, declarative, and compositional than free-form conversations with language models or community evaluation harnesses comparing model capabilities on self-contained tasks. This paradigm necessitates a uniform framework for posing our research questions. For this, we build our testbed in the DSPy framework out of convenience and based on its popularity for optimizing these types of systems, although in principle future declarative languages can support the same research questions we ask about programs, models, and optimizers.

\paragraph{General language programs} Many programs architectures are general: with little to no changes, they can be used for \textit{all} benchmarks. The simplest such program is to use a single DSPy \texttt{Predict} module, which translates to directly calling the language model in a structured manner with the task description and inputs and parsing its outputs. Similarly, we also include a DSPy chain-of-thought (CoT; \citealt{wei2023chainofthoughtpromptingelicitsreasoning}) program that uses CoT prompting, requiring the language model to output both reasoning and answer. We also adapt Archon's modular structure~\cite{saadfalcon2024archonarchitecturesearchframework}  for designing more complex general language programs.
Specifically, we use generators (generate a list of responses given a single query), critics (provide feedback for a list of responses), fusers (compile a list of responses into a single one), and rankers (rank the list of responses).

From these basic building blocks, we build two general language model programs: GenCriticFuser and GenCriticRanker. Namely, both pipelines first employ a generator module to generate a list of responses given inputs from the language model, then a critic module to provide strengths and weaknesses to the list of responses; finally, both a fuser and ranker module is used respectively to get the singular final result.

\paragraph{Specialized programs} We also adopt problem-specific program architectures for certain tasks. For example, we define two retrieval-augmented generation (RAG) programs for some knowledge-intensive and classification tasks.  We define the RAG program with a module that queries the retriever with the task input and then leverages a CoT module to generate the final response with the retrieved documents. Another RAG system, a highly simplified version of Baleen \cite{DBLP:conf/nips/KhattabPZ21}, makes LM calls in a multi-hop design, systematically generating queries for the retriever, performing the retrieval and composing the set of retrieved documents to generate the final answer.

For benchmarks that require interactions within a closed-space environment (agent benchmarks), we build a ReAct \cite{yao2023reactsynergizingreasoningacting} program, which reasons and generates actions simultaneously.

\subsection{Optimizers} \label{subsec:optimizers}

To measure the performances of these compound language programs, it is also important to measure how they perform under various prompt optimization techniques. \tool provides four general prompt optimization techniques for all language model programs.

\textbf{BootstrapFewShot} \cite{DBLP:journals/corr/abs-2212-14024,khattab2024dspy} is a simple heuristic for building few-shot examples for all modules in arbitrary programs. Using a metric and a teacher model, it bootstraps ``successful'' demonstrations from the training set that pass the metric. The process iteratively refines predictions until the maximum number of bootstrapped demonstrations is reached and then included in the final prompt. The \textbf{BootstrapFewShotRandomSearch} optimizer searches through different combinations of bootstrapped few-shot demonstrations using a validation set, applying multiple optimization rounds with random search to select the best-performing set of few-shot examples to include in the final prompt.

\textbf{MIPROv2} \cite{opsahlong2024optimizinginstructionsdemonstrationsmultistage} optimizer produces sets of few-shot examples like BootstrapFewShot and then generates instruction candidates for each predictor in the program using a separate LM proposer program. To search for the best combination of few-shot examples and instructions, the optimizer uses Bayesian Optimization to identify the optimal instruction-examples set for each module in the program.

Finally, as part of this work, we introduce the \textbf{RuleInfer} optimizer, which runs the program through a single iteration of the BootstrapFewShot optimizer, allowing candidates to induce rules based on the successful proposed few-shot demonstrations, tasks, and instructions. These rules are then appended to the original instruction, generating refined instructions directly from the best-selected few-shot demonstrations.

\section{Benchmarking Language Programs}
Evaluating language programs with \tool are simply Cartesian products of dataset tasks, language programs, optimizers, and language models. While some language programs obviously do not work with specific types of datasets, we evaluate all such possible combinations.

To this end, we evaluate 16 distinct tasks from various open-source datasets with six different language models (\textit{gpt-4o}, \textit{gpt-4o-mini}, \textit{o1-mini}, \textit{Llama3.1-8B-Instruct}, \textit{Llama3.2-3B-Instruct}, and \textit{Llama3.3-70B-Instruct}) and a total of more than 10 different language model programs. We also apply four prompt optimization techniques described in \Cref{subsec:optimizers} to all language model programs with different hyperparameters (\Cref{tab:optimizers}). In total, we run over 2000 different combinations of language programs and dataset experiments.

\begin{figure*}[th!]
    \centering
    \includegraphics[width=0.95\linewidth]{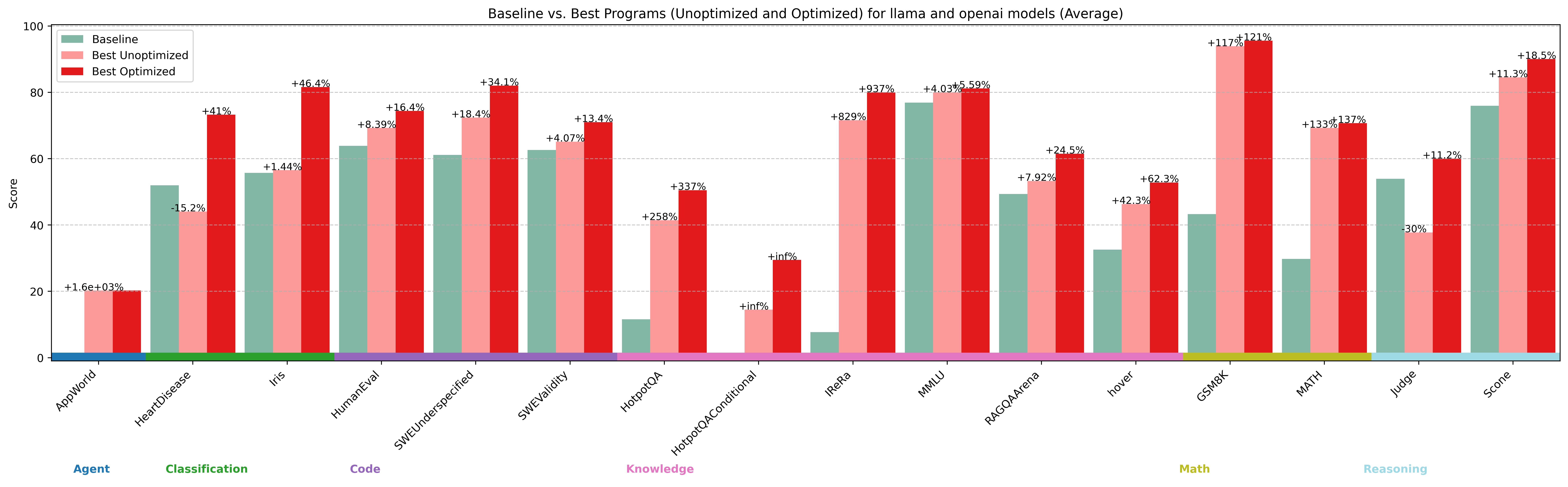}
    \caption{Performance comparison between best-performing programs and the baseline. In most cases, the baseline is a zero-shot call to the LM with task description and task inputs. Scores are averaged from all language models we evaluated, including both OpenAI models and Llama models. \textbf{In almost all tasks, both optimized and unoptimized programs perform better than the raw model prediction baseline.}}
    \label{fig:best_program_unoptimized}
\end{figure*}

\begin{figure*}[th!]
    \centering
    \includegraphics[width=\linewidth]{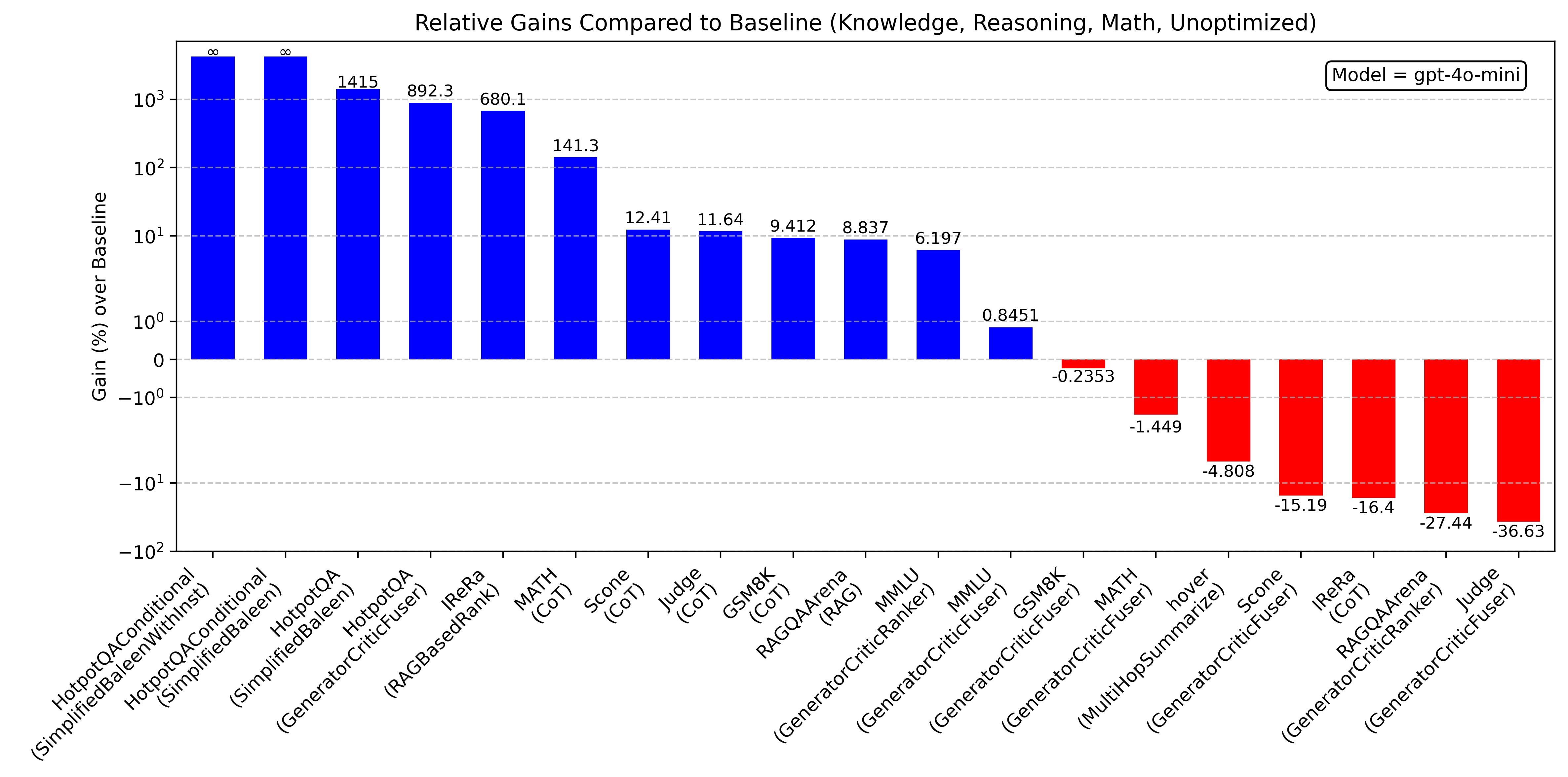}
    \caption{Performance comparison to Baseline, with \textit{best} performing and \textit{worst} performing programs for all Knowledge, Reasoning, and Math tasks. All programs are unoptimized. \textbf{On the same dataset, different language programs' performances vary. Similarly, the same language program's performance varies on different datasets.}}
    \label{fig:program_comparison}
\end{figure*}

\section{Costs of Programs and Optimizers}\label{sec:eval_costs}

In this section, we analyze the cost-performance trade-offs of using language models with varying inference costs and capabilities when combined with language programs and optimizers. Figure~\ref{fig:pareto} visualizes these trade-offs as Pareto curves aggregated over multiple datasets in LangProBe. The x-axis uses a logarithmic scale for inference costs to accommodate the wide range of costs across different configurations. %

We make the following observations from the Pareto curves: 1) \textbf{Every point on the Pareto-optimal curve is instantiable}: The Pareto-curves are calculated as an upper-left convex hull over linearly scaled performance-cost axes, and hence, every point on or below the curves is instantiable by cost-aware load balancing between the configurations at either end of a Pareto segment. 2) \textbf{Use of language programs or optimizers enables achieving better performance at lower cost:} the \textcolor{model_program_optimizer_color}{Model+Program+Optimizer} Pareto curve dominates (achieve better performance at lower cost) the \textcolor{model_program_color}{Model+Program} and \textcolor{model_optimizer_color}{Model+Optimizer} Pareto curves, which in turn dominate the \textcolor{model_color}{Model} Pareto curve, implying that using language programs and optimizing them has the capacity to support significant improvements not only with respect to performance, but also cost. 3) \textbf{Smaller models with effective programs and optimizers outperform larger models, at a lower cost}: We observe that often a smaller or cheaper LM with language program or optimization (or both) outperform configurations with larger or expensive LMs lacking language programs or optimization both in terms of cost and absolute performance. On aggregate, in Figure~\ref{fig:pareto}, we observe that gpt-4o-mini with language programs and optimization achieves 11.68\% higher score than gpt-4o's baseline at just 50\% of the cost, and gets slightly better performance than gpt-4o with language programs at 10\% of the cost.

Among individual datasets, in Figure~\ref{fig:pareto_hotpotqa}, we see that gpt-4o-mini with program composition and optimization achieves 33.2\% better result than gpt-4o without program composition and optimization at 18\% lesser cost. These instances reflect the importance of composing compound systems to prompt language models and optimizing these systems, which can lead to increased downstream task performance at fractional costs.%

In some instances, we observed that using an optimizer can actually reduce the inference cost for the same program while boosting performance. For instance, in GSM8K (\Cref{fig:pareto_gsm8k}), we see that gpt-4o-mini with optimization (BootstrapFewShotWithRandomSearch) achieves 8\% better result at 5\% lesser inference cost than gpt-4o-mini without optimization. Since optimizing the program with BootstrapFewShotWithRandomSearch includes few-shot examples in the prompt, which increases the number of input-context tokens presented to the model but excludes the generation of lengthy reasoning traces in the output, and the notion that LM input token costs are cheaper compared to output token costs, we observe such cases.

In conclusion, our results demonstrate that language programs and optimizers can significantly enhance performance and cost-efficiency across tasks compared to raw model predictions on the aggregate.%

\section{Language Program Design} \label{sec:eval_programs}

We now address the research question of what language program architecture achieves higher quality, e.g. whether structured language programs consistently outperform direct model prediction baselines and which programs contribute the most to performance gains. Additionally, we examine whether the effectiveness of different language programs varies across tasks, highlighting the conditions under which they provide the greatest benefit. 

First, we show an overview of performance comparison between best language programs and raw model prediction baseline in \Cref{fig:best_program_unoptimized}. In almost all cases, using \textbf{some} selected language programs, either unoptimized or optimized, gives a better performance than the raw model prediction baseline.

\subsection{Which language program designs provide
more improvements?}
Although \Cref{fig:best_program_unoptimized} provides a clear overview of the best-performing language programs on each dataset, it does not easily show the relative improvements achieved by different programs. To address this, we present \Cref{fig:program_comparison}, which highlights, for each dataset in the Knowledge, Reasoning, and Math categories, both the best-performing and worst-performing programs and their performance relative to the baseline.

From \Cref{fig:program_comparison}, the performance of different language programs varies significantly when compared to the baseline. On the left side, we observe significant improvement over the baseline by mostly retriever-augmented generation programs. Conversely, other programs, including GeneratorCriticRanker for RAGQAArena and GeneratorCriticFuser for JudgeBench perform worse than the baseline. Inspecting the evaluation traces, we find that when the program involves multiple modules, errors (like wrong format, factual errors, or hallucination) cascade from one module flow into other modules, causing errors to aggregate. Luckily, the error aggregation patterns, especially parsing errors, are mitigated through carefully curated few-shot examples and instructions from different optimization techniques as highlighted by the positive performance with both optimized variances.

Overall, these results demonstrate plenty of gains due to compositional program architectures, but they also show that human judgment about which compositions to pursue (or a well-scoped search for tasks with enough data) is still required for best performance. In other words, there is no general ``set it and forget it'' strategy within the scope we considered.

\subsection{When do language programs provide more improvements?}
From both \Cref{fig:program_comparison} and the individual Pareto curves on each dataset in \Cref{sec:appendix:pareto}, we observe larger improvement from language programs often associated with datasets or tasks that require \textit{additional} information to complete. For example, HotPotQAConditional requires answering questions in a specific format that is unknown to the language model; IReRa is a classification task that needs additional information about the labels, etc. Language programs often show significant improvement over these datasets, which resembles a large class of real-world applications. On the other hand, tasks that language models are trained to perform, like MMLU or HumanEval, often see little to no benefit from unoptimized language programs. \Cref{fig:pareto_humaneval} plots an overlapping Pareto curve for \textcolor{model_program_color}{Model+Program} and the baseline, demonstrating no performance benefits from using unoptimized language programs for HumanEval.

\begin{figure}[ht!]
    \centering
    \includegraphics[width=\linewidth]{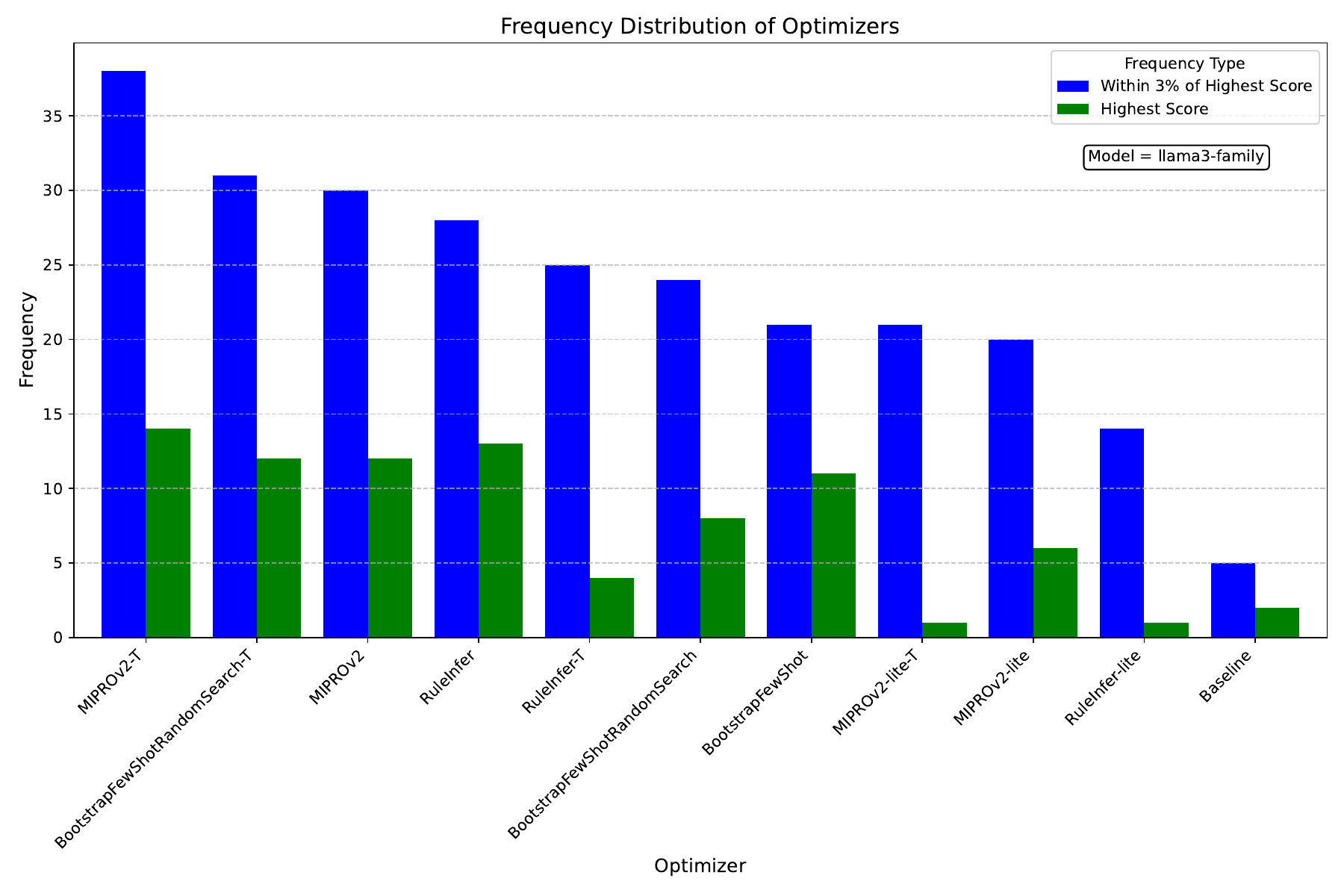}
    \caption{Frequency Distribution for individual optimizer performance, ranked by the number of times that an optimizer applied on a program is within 3\% of that program's highest score (blue bar). We also note the number of the highest-performing optimizer as the top score (green bar). From the plot, \textbf{MIPROv2-T, which uses a stronger model for optimization to propose better instructions combined with corresponding few-shot examples through Bayesian search, works the best.}}
    \label{fig:optimizer_perf}
\end{figure}

\begin{figure*}[th!]
    \centering
    \includegraphics[width=\linewidth]{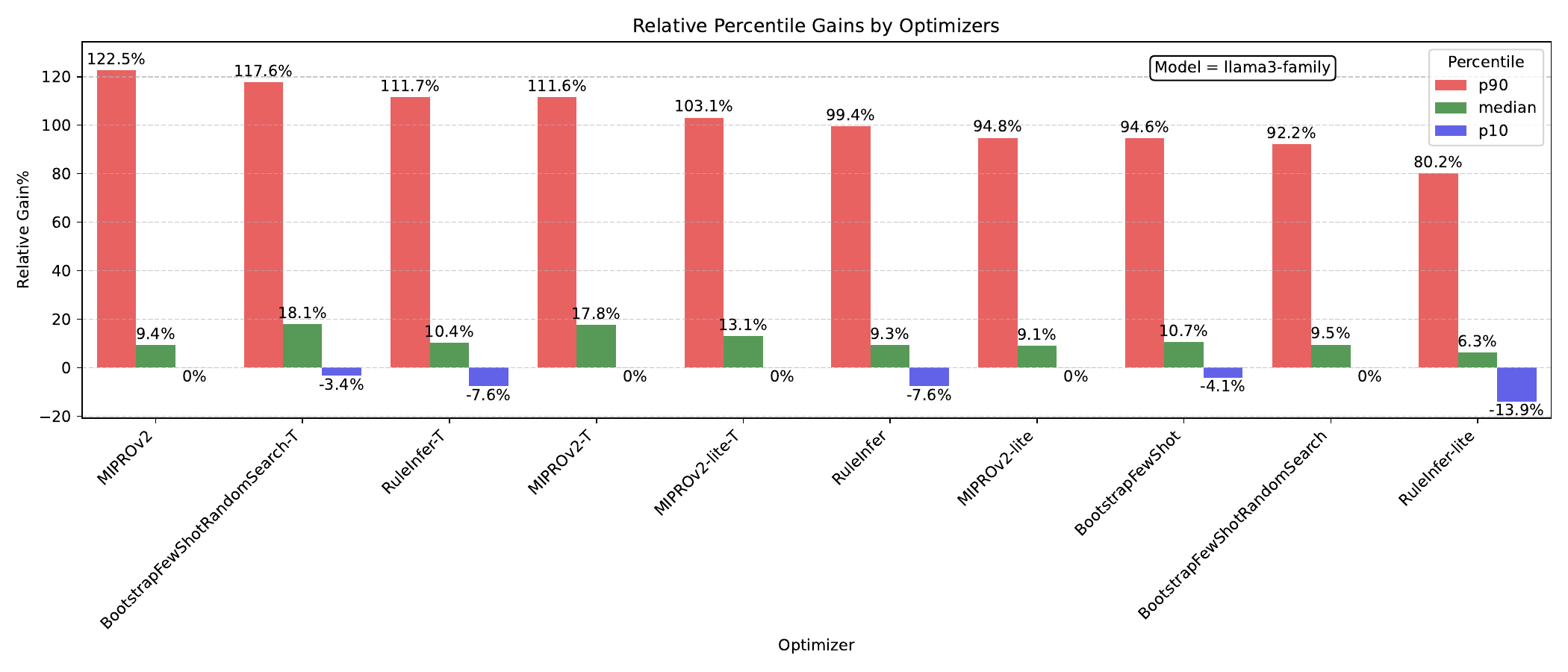}
    \caption{For all the experiments with Llama models, the relative gains by the optimizer compared to the same unoptimized program. We report 90th percentile, median, and 10th percentile. Optimizers with "lite" suffixes are configured with less compute resources, and optimizers with "T" are configured with a stronger optimizer model (gpt-4o-mini). From this plot, \textbf{in best case scenarios, all optimizers provide large performance gains of more than 80\%. The median performance gains are from 6.3\% to 18.1\%. However, in some cases, optimizers do degrade the performance of the unoptimized program.}}
    \label{fig:optimizer_relative_gain}
\end{figure*}

\section{Optimizers} \label{sec:eval_optimizers}

As in \Cref{sec:eval_programs}, programs are important for most datasets, as they provide additional context or more opportunities to reason with the language model. In addition to program design and architecture, prompt-based optimizations are also crucial to the performance of a compound AI system. In this section, we discuss whether the existing optimization techniques are effective by answering the following research questions.

\subsection{What optimization techniques are more effective than others?}

We rank the performance of each individual optimizer in \Cref{fig:optimizer_perf}. From the plot, we observe that general optimizers that perform optimal instruction finding (MIPROv2 and its variants) and in-context learning (BootstrapFewShotRandomSearch and its variants) are leading in general, both having more experiments that appear within 3\% of the highest score for the same dataset and program among all optimizer settings. The rule induction optimizers "RuleInfer" are good at obtaining the best scores but perform slightly worse to generalize for more tasks. This is due to its specialized nature, which allows RuleInfer to only work for tasks with obvious and conclusive rules from the examples.

Additionally, optimizers using a stronger model internally (suffices with "T") generally perform better with an exception on RuleInfer. Because MIPRO and BootstrapFewshot all depend on good few-shot demonstrations, a stronger internal model provides better performance by generating better traces at a relatively lower cost.

\subsection{How do performance gains from different
optimizations vary across the distribution?}

We list the relative gain for all optimizers in \Cref{fig:optimizer_relative_gain}. From the plot, at the 90th percentile, all optimizers demonstrate strong performance, ranging from 80.2\% up to 122.5\%. This illustrates that if used carefully, optimizers can provide substantial performance gains. For example, for the HeartDisease dataset, the MIPROv2 optimizer finds instructions like ``\textit{using the provided patient information, predict whether the patient has heart disease by analyzing the clinical parameters step by step and providing a rationale for your conclusion}'' and a few good few-shot examples with detailed LM-generated reasoning. These optimized prompts increase the performance for Llama-3.2-3B from 26.32 to 76.32, resulting in an 189.97\% increase.

The median of performance gains suggests the common effect of using optimizers. For most tasks, we expect a 5\% to 20\% from the optimizers. Finally, in some rare cases, performance degradation happens. Because most optimizers employ a separate validation set to control the quality of optimizations, the optimized language program may overfit the validation set, causing a performance drop on the final test set. RuleInfer and its variants have the most performance degradation due to non-transferable rules. Although the rules induced by the optimizer work well for the validation set, some may not directly help with the test set performance.

\section{Conclusion}

We presented \tool, a benchmark for evaluating language programs across a range of tasks. We used it to investigate far more comprehensively than prior work the interactions between architectures, optimization strategies, language models, and the resulting quality and cost tradeoffs. We hope that this starts a line of work on studying this new category of AI systems and that our current findings can already begin to offer practical guidance for researchers and practitioners designing and optimizing modular AI systems. Lastly,  one of the primary goals of \tool is to facilitate the development and comparison of new language program architectures and optimization strategies.

\section*{Limitations}
Due to compute constraints, \tool is not able to include more language programs (e.g., program-of-thought) and datasets (e.g., SWE-bench), which could allow more insightful observations. \tool does not run full evaluation on the newest reasoning models, like DeepSeek-R1 or OpenAI o3-mini, also due to budget and time constraint.

\bibliography{reference}

\appendix
\newpage

\section{Dataset Descriptions} \label{sec:appendix:bench}
\tool uses the following open-sourced datasets for research purposes only. \\

\noindent
\textbf{AppWorld} \cite{DBLP:conf/acl/TrivediKHMDLGSB24}

\noindent
\underline{Task}: AppWorld databases start in some initial state set by the benchmark. The agent should take certain actions during the execution to change the database into another state.

\noindent
\underline{Input}: A question about mobile applications, along with its supervisor's information like name, phone number, and email.

\noindent
\underline{Output}: python code that will be executed by AppWorld server.

\noindent
\underline{Evaluation metrics}: the output python code will be executed by the AppWorld server and check if the database after execution is the same as the ideal final state. Success is marked by passing all the unit tests for state check and no unwanted actions are taken. Otherwise, it's a failure. 

\noindent
\underline{License}: Apache 2.0 \\

\noindent
\textbf{SweBenchVerifiedAnnotation} \cite{SWEbench}

\noindent
\underline{Task}: to examine each sample in the SWE-bench test set for properly defined issue descriptions and unit tests with appropriate scope by giving them scores.

\noindent
\underline{Input}: the repository name containing the issue, the issue description, gold patch, test patch,  and the names of the tests in the test patch that will be used to evaluate the solution.

\noindent
\underline{Output}: a score from 0 to 3 based on certain 4 criteria.

\noindent
\underline{Evaluation metrics}: the ground truth is also a score based on the same criteria. We calculate the string equivalence between the predicted score and the ground truth score. 

\noindent
\underline{License}: MIT\\

\noindent
\textbf{MATH} \cite{DBLP:conf/nips/HendrycksBKABTS21}

\noindent
\underline{Task}: The Mathematics Aptitude Test of Heuristics (MATH) dataset consists of problems from mathematics competitions, including the AMC 10, AMC 12, AIME, and more. Each problem in MATH has a full step-by-step solution, which can be used to teach models to generate answer derivations and explanations.

\noindent
\underline{Input}: a math question written in LaTeX and natural language.

\noindent
\underline{Ground truth}: step-by-step solution written in LaTeX and natural language with the final answer enclosed in LaTeX's \textbackslash boxed tag.

\noindent
\underline{Evaluation metrics}: string equivalence between the predicted mathematical answer and the gold solution extracted from the LaTeX's \textbackslash boxed tag. We adopt \citet{DBLP:conf/nips/HendrycksBKABTS21}'s evaluation. Arguably, a more prominent evaluator like Math-Verify \cite{Kydlicek2025mathverify} would report fairer (and higher) scores for all programs and optimizers, which is left as future work.

\noindent
\underline{License}: MIT\\

\noindent
\textbf{GSM8K} \cite{cobbe2021trainingverifierssolvemath}

\noindent
\underline{Task}: solving high-school-level mathematics problems across various topics. These problems are presented in natural language, and the model is required to produce correct solutions. 

\noindent
\underline{Input}: The question to a grade school math problem in natural language

\noindent
\underline{Output}: The solution to this problem with step by step reasoning in natural language. The numerical solution is always at the end of the solution string.

\noindent
\underline{Evaluation metrics}: the ground truth is the solution to this problem with step-by-step reasoning in natural language. We calculate the integer equivalence between the predicted mathematical answer and the gold solution. 

\noindent
\underline{License}: MIT\\

\noindent
\textbf{HotPotQA} \cite{DBLP:conf/emnlp/Yang0ZBCSM18}

\noindent
\underline{Task}: HotpotQA is a new dataset with 113k Wikipedia-based question-answer pairs. Our task is to answer questions that require reasoning across multiple supporting documents.

\noindent
\underline{Input}: question in natural language

\noindent
\underline{Output}: answer to the question in natural language

\noindent
\underline{Evaluation metrics}: ground truth is the answer to the question in natural language. We evaluate the string equivalence between the predicted answer and the ground truth answer. 

\noindent
\underline{License}: CC BY-SA 4.0 \\

\noindent
\textbf{HumanEval} \cite{chen2021evaluatinglargelanguagemodels}

\noindent
\underline{Task}: The HumanEval dataset released by OpenAI includes 164 programming problems with a function signature, docstring, body, and several unit tests. The task is to produce reliable and executable code that passes the unit tests given.

\noindent
\underline{Input}: the prompt for function specification that includes necessary import statements, function signatures, and docstring of unit tests.

\noindent
\underline{Ground truth}: generated code snippet following the function specification.

\noindent
\underline{Evaluation metrics}: binary indicator that specifies whether generated code passes the test cases defined in the ground truth. 

\noindent
\underline{License}: MIT\\

\noindent
\textbf{MMLU} \cite{hendrycks2021measuringmassivemultitasklanguage}

\noindent
\underline{Task}: answer multiple-choice questions across a broad range of knowledge domains.

\noindent
\underline{Input}: description of the question along with its four options in natural language.

\noindent
\underline{Ground truth}: the correct option for this question

\noindent
\underline{Evaluation metrics}: ground truth is provided as the correct option in natural language (“A”, “B”, “C”, or “D”). We evaluate the string equivalence between the predicted option and the ground truth option.

\noindent
\underline{License}: MIT\\

\noindent
\textbf{IReRa} \cite{doosterlinck2024incontextlearningextrememultilabel}

\noindent
\underline{Task}: solve multi-label classification tasks with an extreme number of classes

\noindent
\underline{Inputs}: a textual description

\noindent
\underline{Outputs}: all the ESCO job skill labels mentioned in natural language.

\noindent
\underline{Evaluation metrics}:  rank-precision (RP) of the produced rankings, calculated from a list of true relevant items and a list containing the predicted items in ranked order.

\noindent
\underline{License}: MIT\\

\noindent
\textbf{Heart Disease} \cite{heart_disease_45}

\noindent
\underline{Task}: classify the patient’s heart disease status based on the information provided

\noindent
\underline{Inputs}: textual description of patient heart disease attributes, including age, slope, chol, ca, thal, restecg, exang, trestbps, cp, thalach, fbs, oldpeak, and sex.

\noindent
\underline{Outputs}: A binary classification ("yes" or "no") indicating whether the patient has heart disease.

\noindent
\underline{Evaluation metrics}: we evaluate the string equivalence between the predicted diagnosis and the ground truth diagnosis (“yes” or “no”).

\noindent
\underline{License}: CC BY 4.0\\

\noindent
\textbf{HoVer} \cite{jiang2020hoverdatasetmanyhopfact}

\noindent
\underline{Task}: perform multi-hop evidence retrieval of a claim to determine whether it’s supported or not.

\noindent
\underline{Inputs}: a claim about a fact in natural language. 

\noindent
\underline{Outputs}: documents being retrieved concatenated as a single string.

\noindent
\underline{Evaluation metrics}: The ground truth is the fact the model is required to retrieve. We calculate the proportion of examples in which the model is required to retrieve at least one supporting fact from each supporting document and accurately predict the correct label.

\noindent
\underline{License}: CC BY-SA 4.0\\

\noindent
\textbf{Iris} \cite{iris_53}

\noindent
\underline{Task}: predict the species of an iris flower based on its sepal and petal measurements.

\noindent
\underline{Inputs}: petal and sepal dimensions in cm about the iris species in natural language.

\noindent
\underline{Outputs}: the class of iris plant (setosa, versicolor, or virginica).

\noindent
\underline{Evaluation metrics}: the ground truth is the iris class in natural language. We evaluate the string equivalence between the predicted class and the ground truth class. 

\noindent
\underline{License}: CC BY 4.0\\

\noindent
\textbf{Scone (Scoped Negation Benchmark)} \cite{She_2023}

\noindent
\underline{Task}:  ScoNe-NLI contains contrast sets of six examples where entailment relations are impacted by the scope of one or two negations. The main task is to test how well models understand and reason about negation in natural language.

\noindent
\underline{Inputs}: a context dependent question and its context in natural language

\noindent
\underline{Outputs}: “yes” or “no”

\noindent
\underline{Evaluation metrics}: ground truth answer to the question is “yes” or “no”. We evaluate the string equivalence between the predicted answer and the ground truth answer.

\noindent
\underline{License}: CC0 1.0 \\

\noindent
\textbf{RAG-QA Arena "Technology"} \cite{han2024ragqaarenaevaluatingdomain}

\noindent
\underline{Task}: generate quality answers for technology questions. 

\noindent
\underline{Inputs}: question in natural language.

\noindent
\underline{Outputs}: response to the question in natural language.

\noindent
\underline{Evaluation metrics}: ground truth is a response in natural language. We calculate the semantic F1 score between the model response and the ground truth response. Note here the language model used to evaluate F1 is the same as the model being evaluated.

\noindent
\underline{License}: Apache 2.0\\

\noindent
\textbf{JudgeBench} \cite{tan2024judgebenchbenchmarkevaluatingllmbased}

\noindent
\underline{Task}: evaluating LLM-based judges for objective correctness on challenging response pairs. 

\noindent
\underline{Inputs}: a question, response A and response B in natural language

\noindent
\underline{Outputs}: which response is better. Either  A>B or B>A.

\noindent
\underline{Evaluation metrics}: ground truth is which one is better, either “A>B” or “B>A”.  string equivalence between the predicted answer and the ground truth answer.

\noindent
\underline{License}: MIT\\

\section{Language Program Descriptions} \label{sec:appendix:language_program}

\noindent
\textbf{CoT}

\noindent
Given an instruction prompt, Chain of thought produces a step-by-step explanation leading to the final answer. 

\noindent
\underline{Number of LLM calls}: 1 \\

\noindent
\textbf{RAG}

\noindent
Retrieval Augmented Generation first retrieves top-k most relevant passages using a retriever. These retrieved passages are then integrated as context into the model’s input and passed through a Chain-of-Thought reasoning process to generates a final response.

\noindent
\underline{Number of LLM calls}: 1 \\

\noindent
\textbf{ReActBaseline}

\noindent
Re-Act is a combination of reasoning and action in a step-by-step manner. Usually, the action involves retrieving relevant information from external sources and then integrates reasoning and action-based decision-making into the model’s process. In our case of AppWorld, the action involves generating code that will be executed by the AppWorld server.

\noindent
\underline{Number of LLM calls}: equivalent to the number of reasoning steps (the number of reasoning steps defaults to 40 in the case of AppWorld).\\

\noindent
\textbf{ReActAugmented}

\noindent
Same as ReActBaseline but with few-shot demonstration added. 

\noindent
\underline{Number of LLM calls}: same as ReActBaseline\\

\noindent
\textbf{SimplifiedBaleen}

\noindent
First generates multiple search queries and iterate through each one of them in multiple "hops," where each hop retrieves relevant passages from a knowledge base using a retriever. These retrieved passages are then aggregated as contexts and are passed through a Chain-of-Thought reasoning process to generate the final response.

\noindent
\underline{Number of LLM calls}: equivalent to the number of hops (defaults to be 2).\\

\noindent
\textbf{SimplifiedBaleenWithInst}
\noindent
The setup is the same as SimplifiedBaleen except for a manually written prompt. In our case of HotpotQA conditional, the prompt is the following: When the answer is a person, respond entirely in lowercase.  When the answer is a place, ensure your response contains no punctuation.  When the answer is a date, end your response with “Peace!”. Never end your response with "Peace!" under other circumstances.  When the answer is none of the above categories respond in all caps.

\noindent
\underline{Number of LLM calls}: same as SimplifiedBaleen.\\

\noindent
\textbf{GeneratorCriticRanker}

\noindent
Produce a list of candidate responses from a given instruction prompt, then for each of these responses, it identifies the strengths/weaknesses for each candidate response. Then it returns a ranked list of top-K candidate responses.

\noindent
\underline{Number of LLM calls}: equivalent to the number of candidate * 2 + 1 (the number of candidates defaults to be 5).\\

\noindent
\textbf{GeneratorCriticFuser}

\noindent
Produce a list of candidate responses from a given instruction prompt, then for each of these responses, it identifies the strengths/weaknesses for each candidate response. Then it returns a single, high-quality response. 

\noindent
\underline{Number of LLM calls}: equivalent to the number of candidate * 2 + 1 (the number of candidates defaults to be 5).\\

\noindent
\textbf{RAGBasedRank}

\noindent
It's an in-context learning framework designed for multi-label classification with an extremely large number of classes (IReRa). The process begins by generating queries based on the input and retrieving documents from a fixed retriever. Then the retrieved documents are re-ranked by an LM.

\noindent
\underline{Number of LLM calls}: 2. One for generating the query for retrieval and one for re-ranking the top k retrieved documents. \\

\noindent
\textbf{MultiHopSummarize}

\noindent
It begins by retrieving and summarizing the top k relevant passages for a given claim. Subsequent hops generate refined queries based on previous summaries, retrieving additional passages. In total of 3 iteration performed to gather information about the initial question. 

\noindent
\underline{Number of LLM calls}: equivalent to the number of hops (the number of hops defaults to 7 in the case of HoVer).\\

\noindent
\textbf{CoTBasedVote}

\noindent
It first applies multiple Chain-of-Thought classifiers to generate independent predictions (votes) that includes both the reasoning and the answer to the question. These votes are then assessed and consolidated through critical evaluation of these "opinions". 

\noindent
\underline{Number of LLM calls}: equivalent to the number of voters + 1. (the number of voters defaults to be 3 in the case of Heart Disease).\\

\section{Cost-Performance Pareto Curves for all benchmarks}\label{sec:appendix:pareto}
\begin{figure}[t!]
    \begin{subfigure}{0.99\linewidth}
        \includegraphics[width=\linewidth]{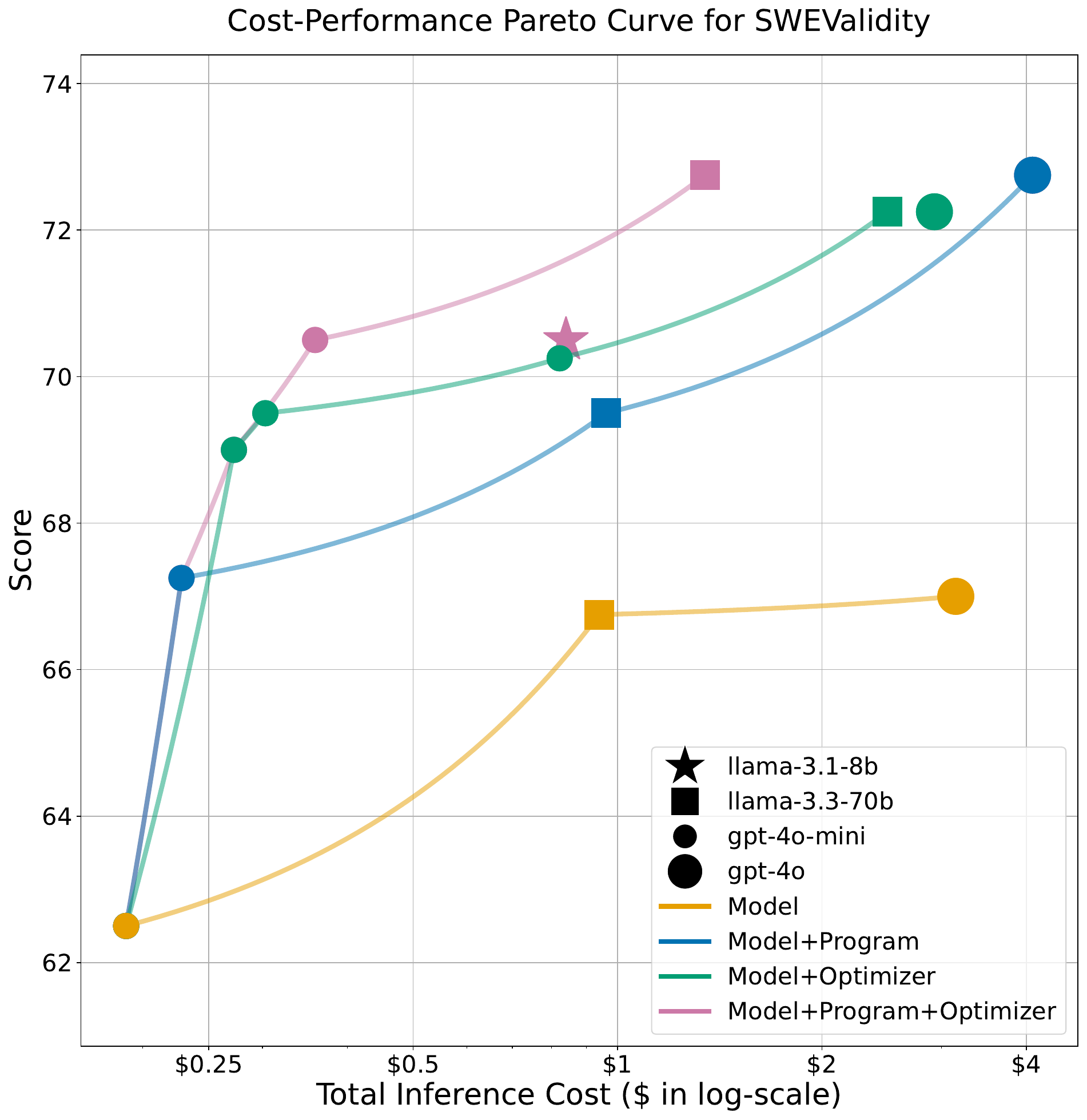}
        \caption{SWEValidity}
        \label{fig:pareto_swevalidity}
    \end{subfigure}\\
    \begin{subfigure}{0.99\linewidth}
        \includegraphics[width=\linewidth]{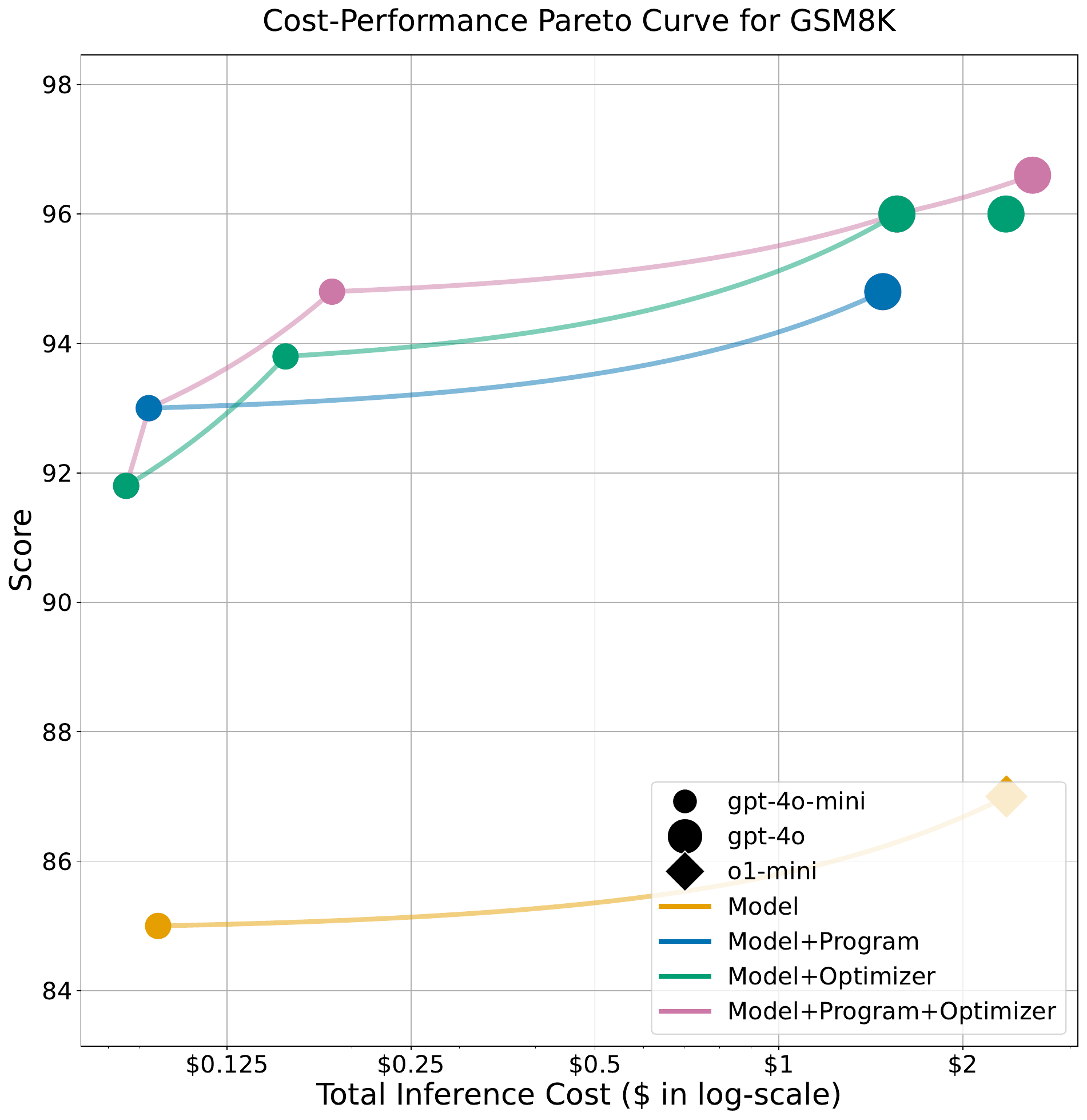}
        \caption{GSM8K}
        \label{fig:pareto_gsm8k}
    \end{subfigure}
    \caption{Performance (Y) vs. Cost (X) Graph for different Benchmarks, Language Programs and Optimizers}\label{fig:pareto_indv3}

\end{figure}
\begin{figure*}[t]
    \centering
    \begin{subfigure}{0.45\linewidth}
        \includegraphics[width=\linewidth]{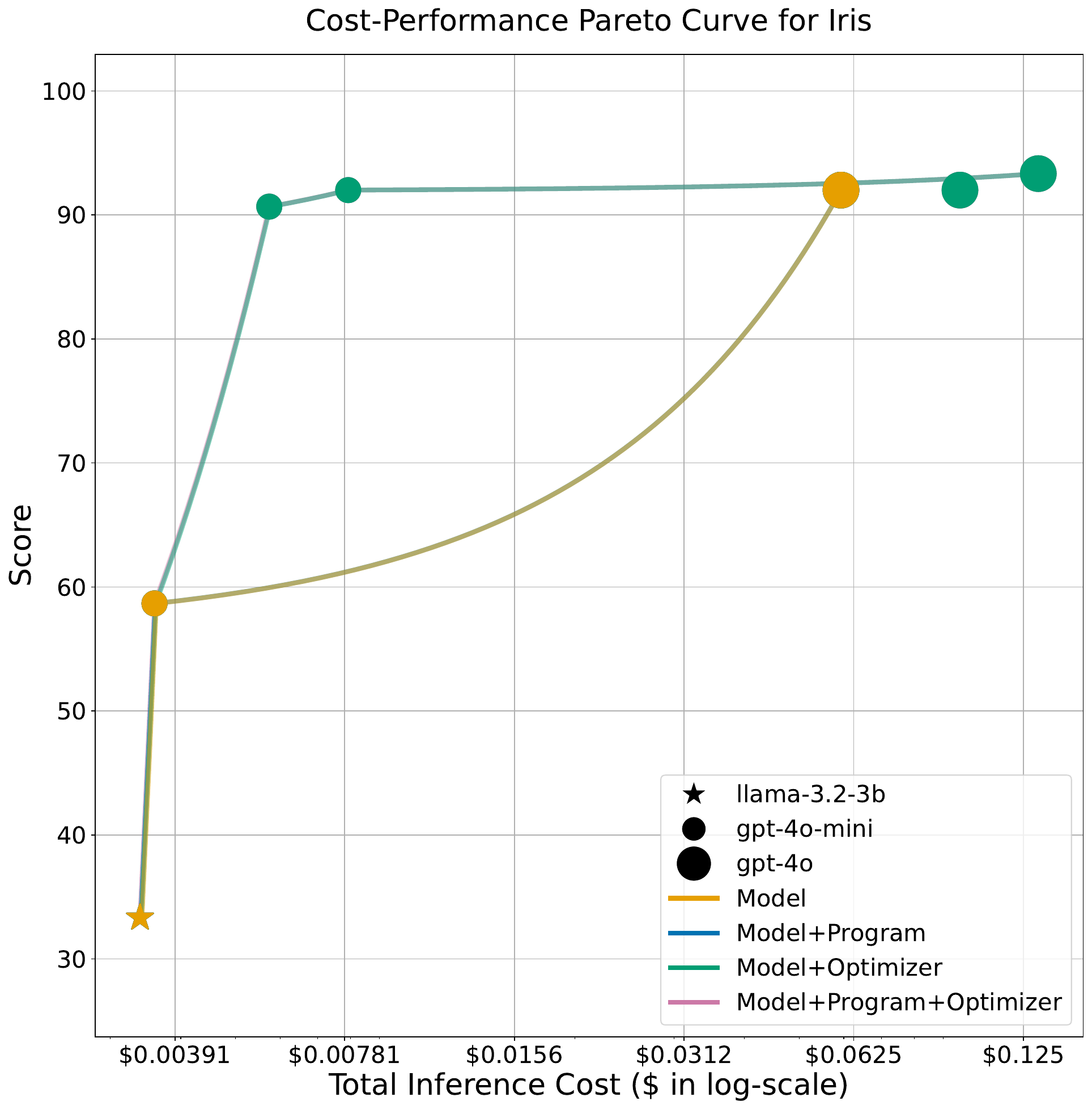}
        \caption{Iris}
        \label{fig:pareto_iris}
    \end{subfigure}
    \begin{subfigure}{0.45\linewidth}
        \includegraphics[width=\linewidth]{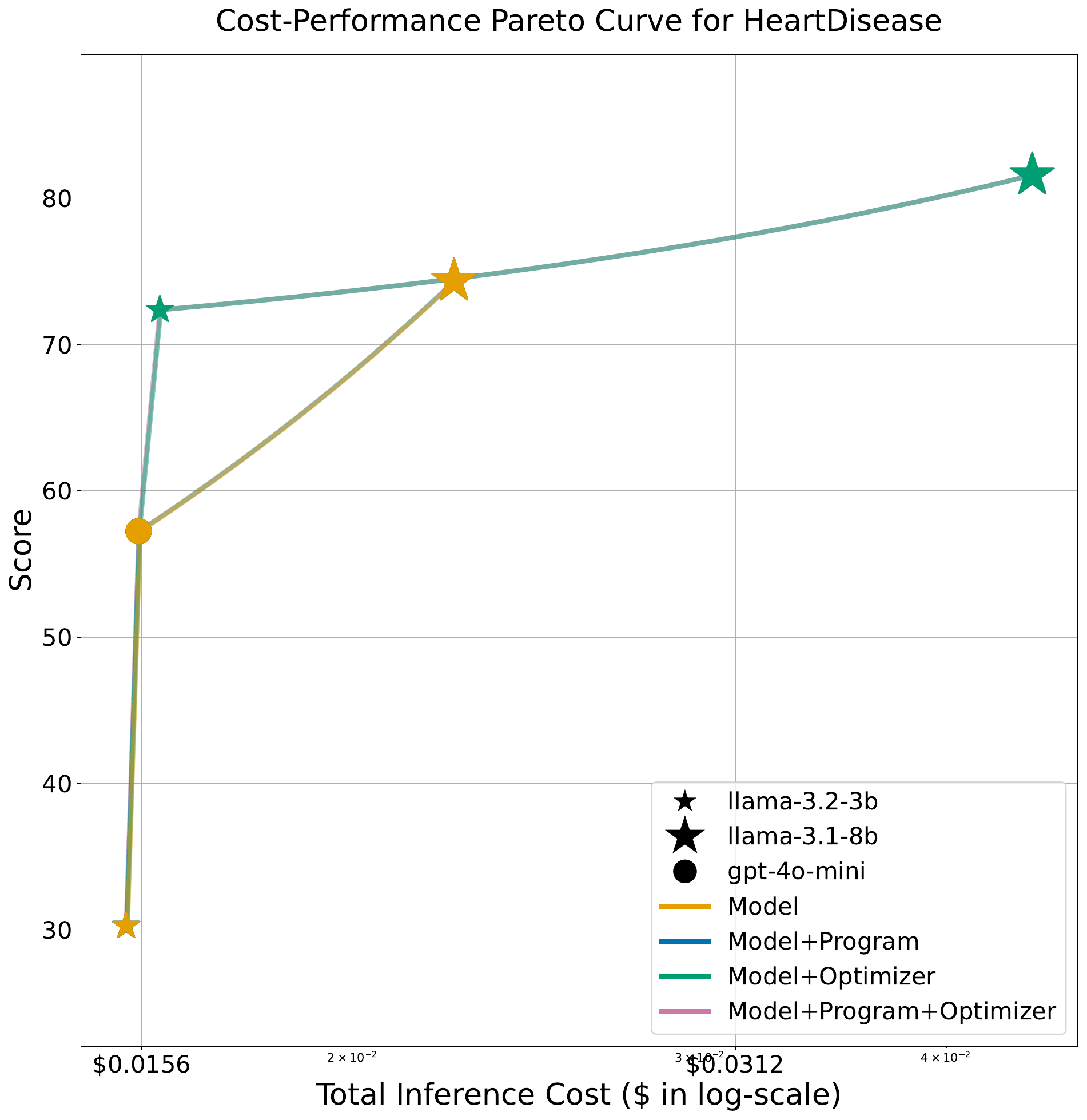}
        \caption{HeartDisease}
        \label{fig:pareto_heartdisease}
    \end{subfigure}\\
    \begin{subfigure}{0.45\linewidth}
        \includegraphics[width=\linewidth]{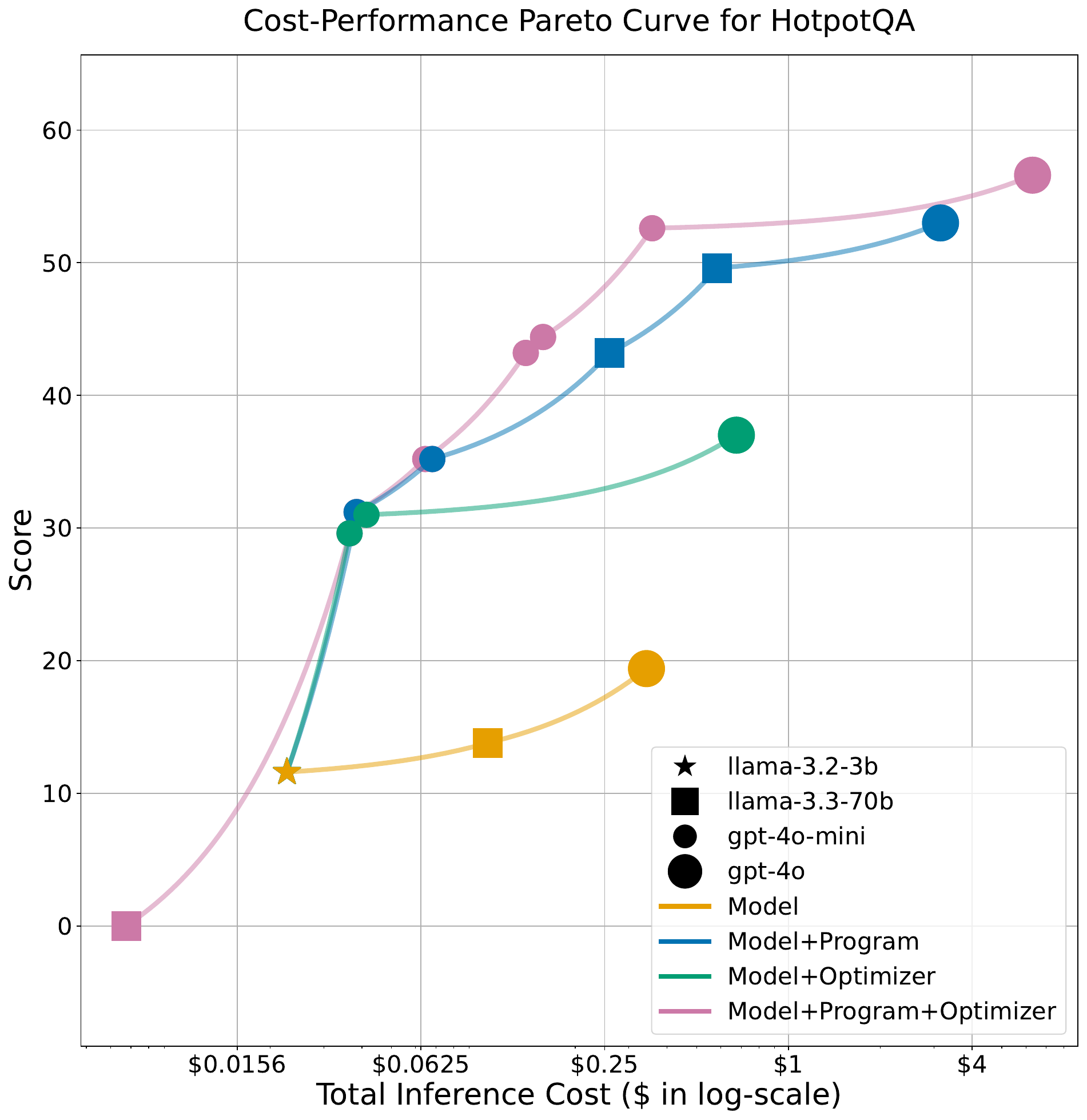}
        \caption{HotpotQA}
        \label{fig:pareto_hotpotqa}
    \end{subfigure}
    \begin{subfigure}{0.45\linewidth}
        \includegraphics[width=\linewidth]{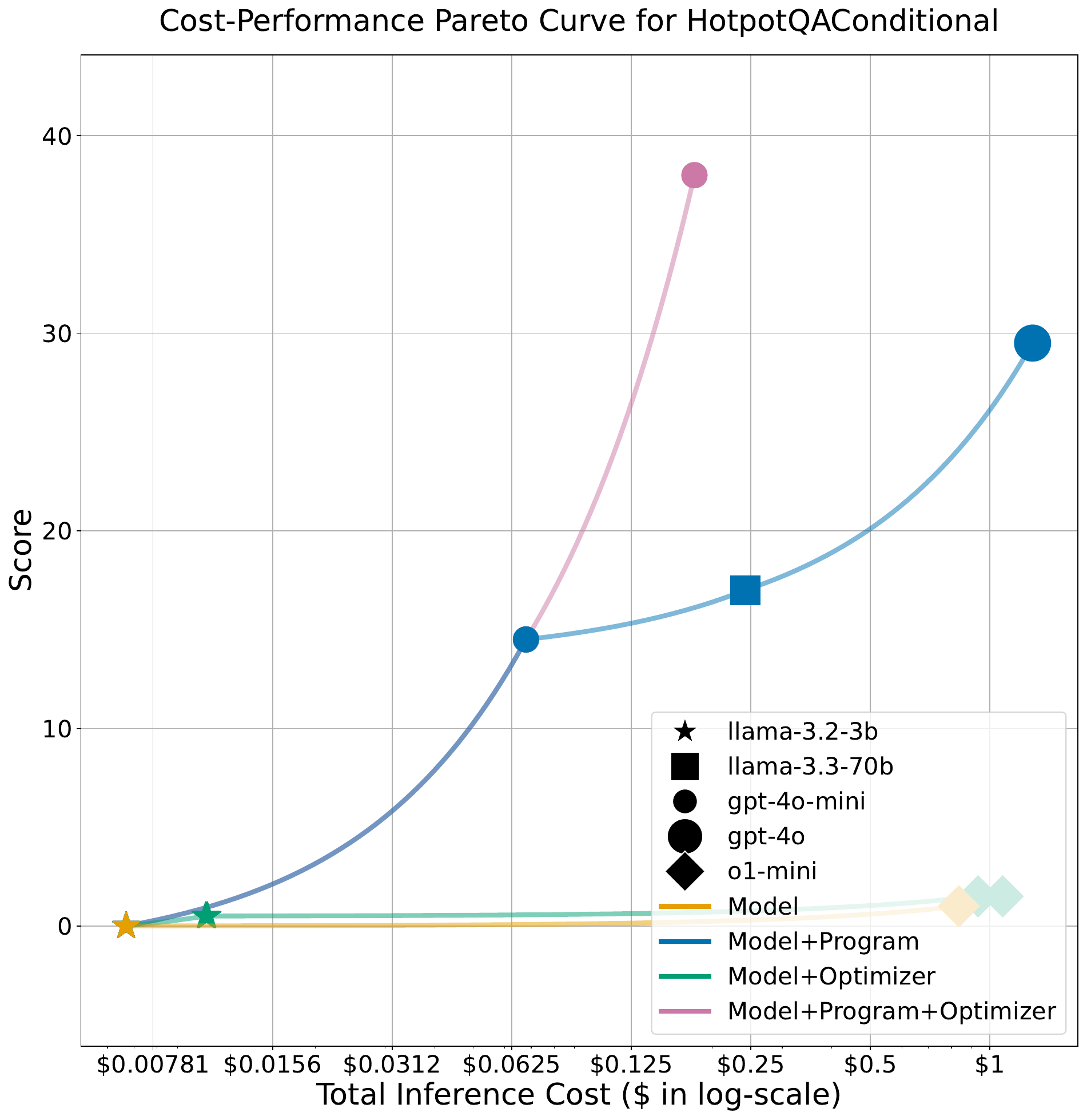}
        \caption{HotpotQAConditional}
        \label{fig:pareto_hotpotqaconditional}
    \end{subfigure}\\
    \begin{subfigure}{0.45\linewidth}
        \includegraphics[width=\linewidth]{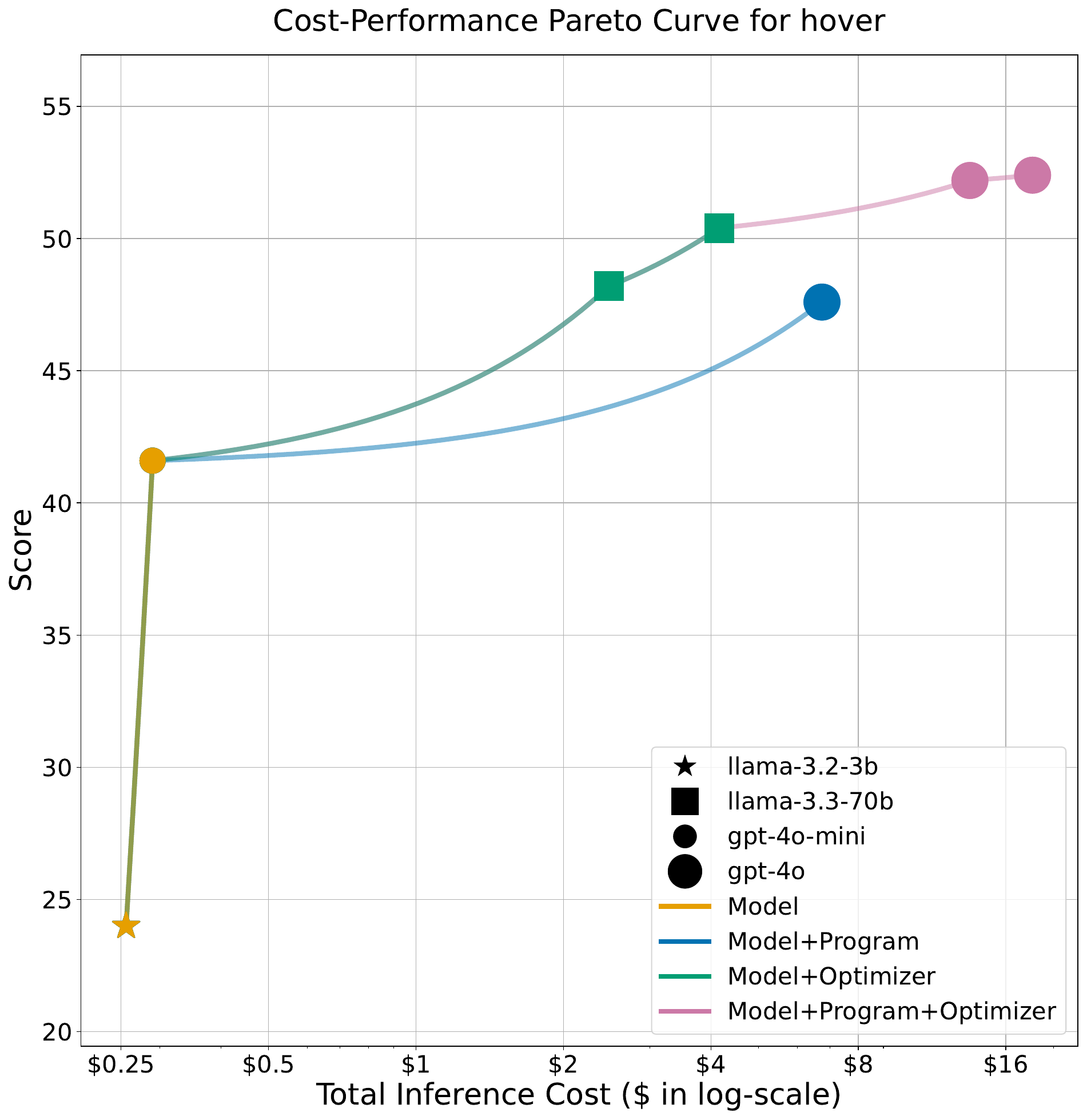}
        \caption{HoVer}
        \label{fig:pareto_hover}
    \end{subfigure}
    \begin{subfigure}{0.45\linewidth}
        \includegraphics[width=\linewidth]{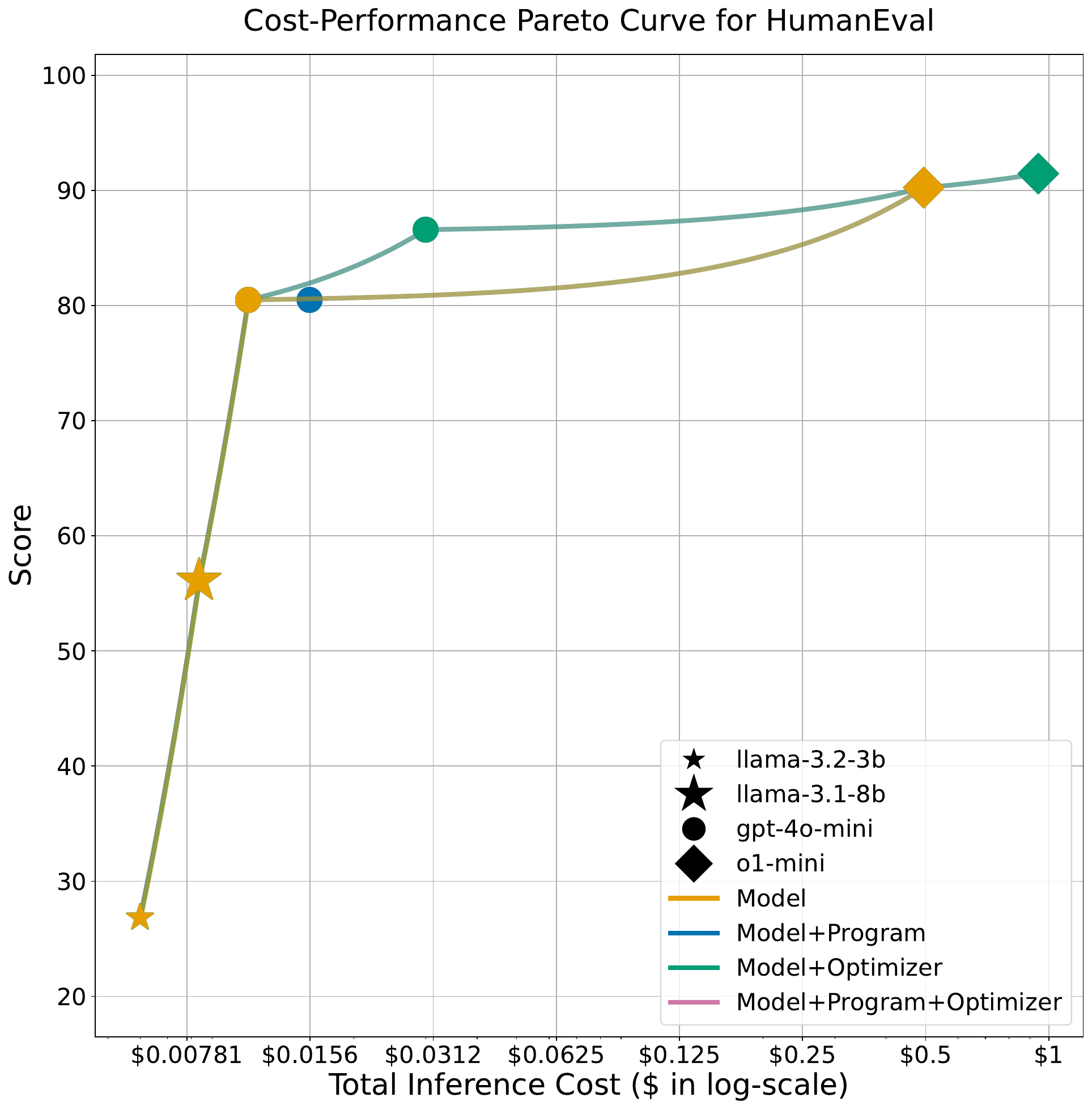}
        \caption{HumanEval}
        \label{fig:pareto_humaneval}
    \end{subfigure}
    \caption{Performance (Y) vs. Cost (X) Graph for different Benchmarks, Language Programs and Optimizers}\label{fig:pareto_indv1}
\end{figure*}
\begin{figure*}
    \begin{subfigure}{0.45\linewidth}
        \includegraphics[width=\linewidth]{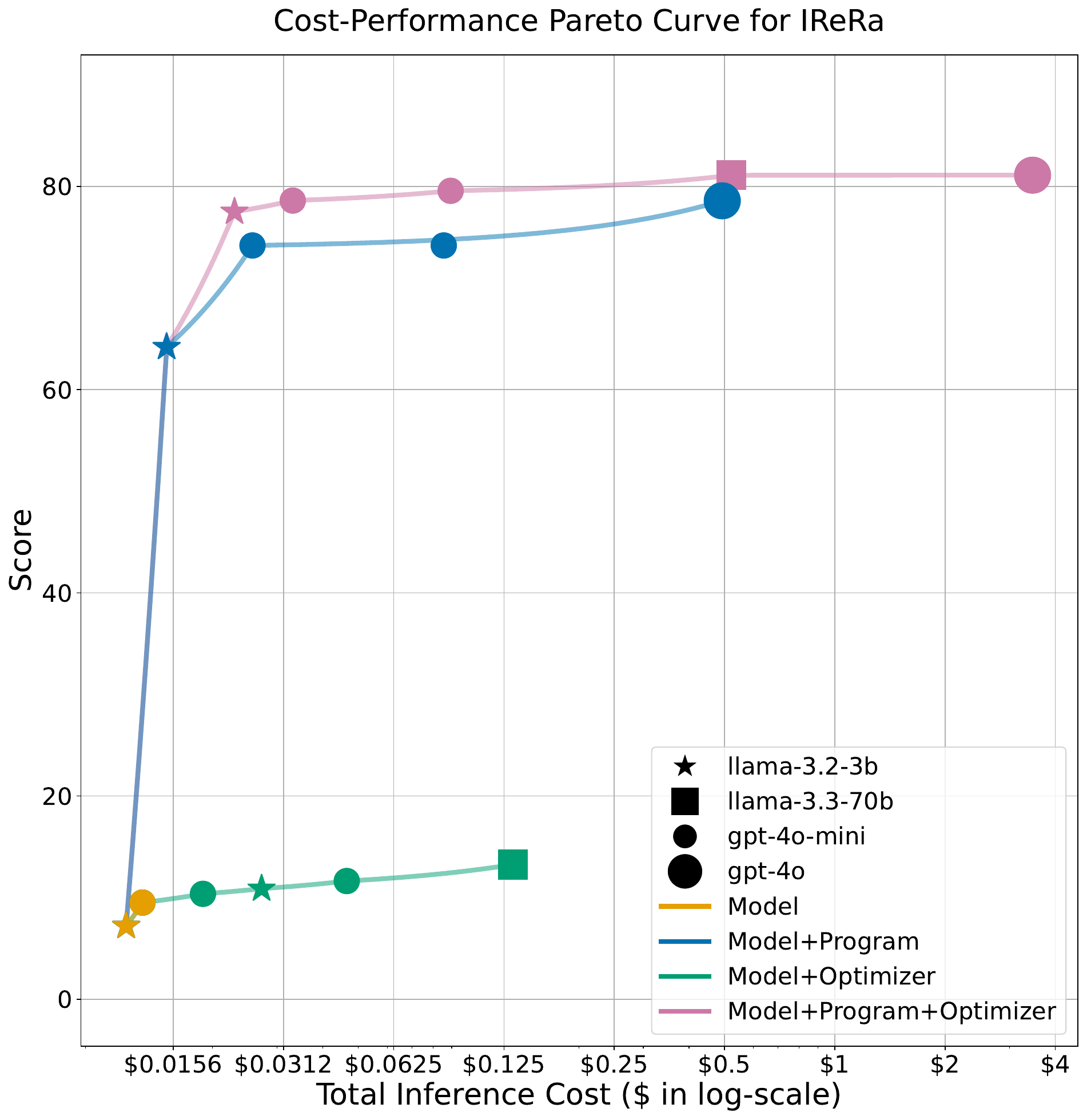}
        \caption{IReRa}
        \label{fig:pareto_irera}
    \end{subfigure}
    \begin{subfigure}{0.45\linewidth}
        \includegraphics[width=\linewidth]{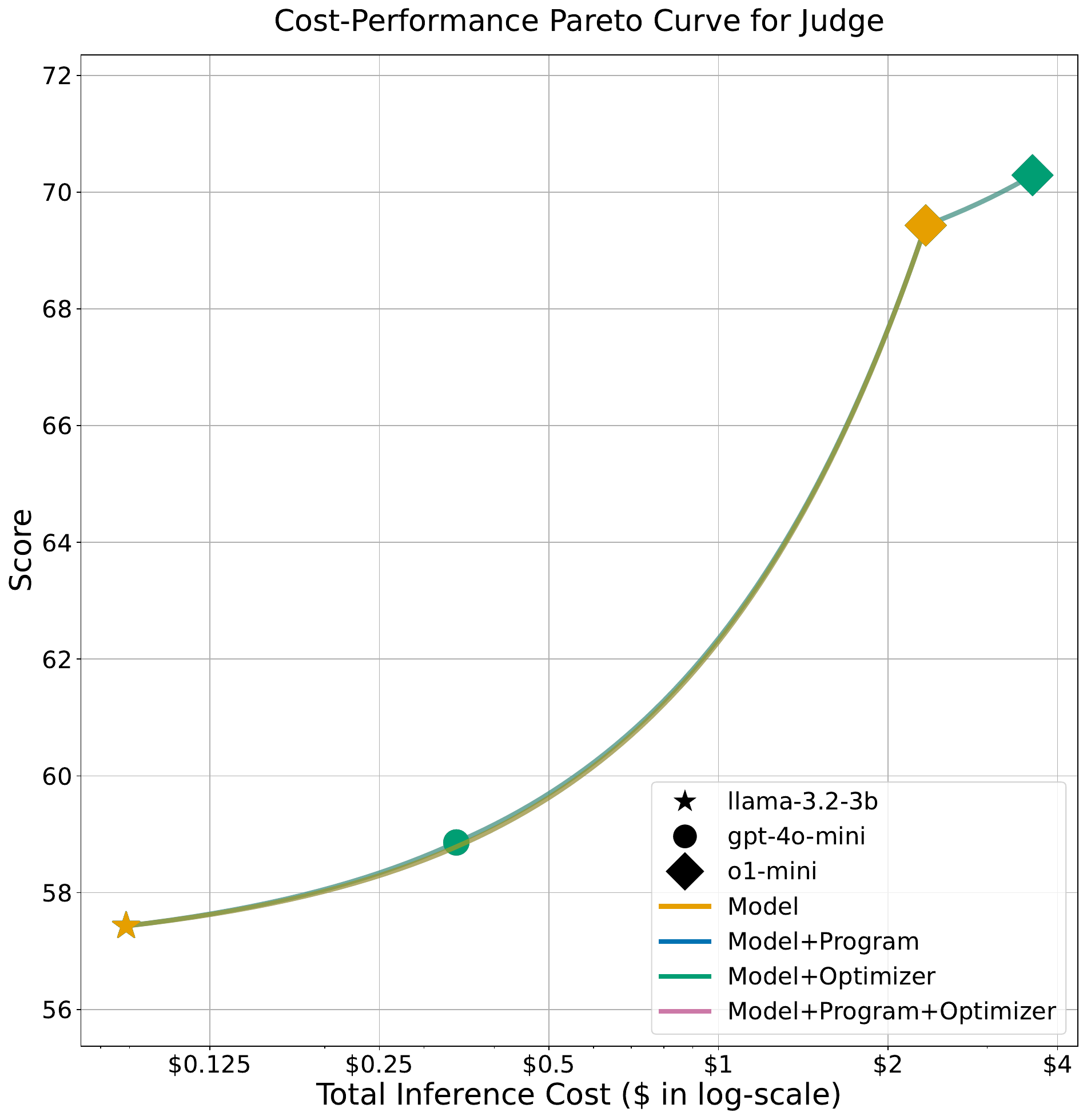}
        \caption{Judge}
        \label{fig:pareto_judge}
    \end{subfigure}\\
    \begin{subfigure}{0.45\linewidth}
        \includegraphics[width=\linewidth]{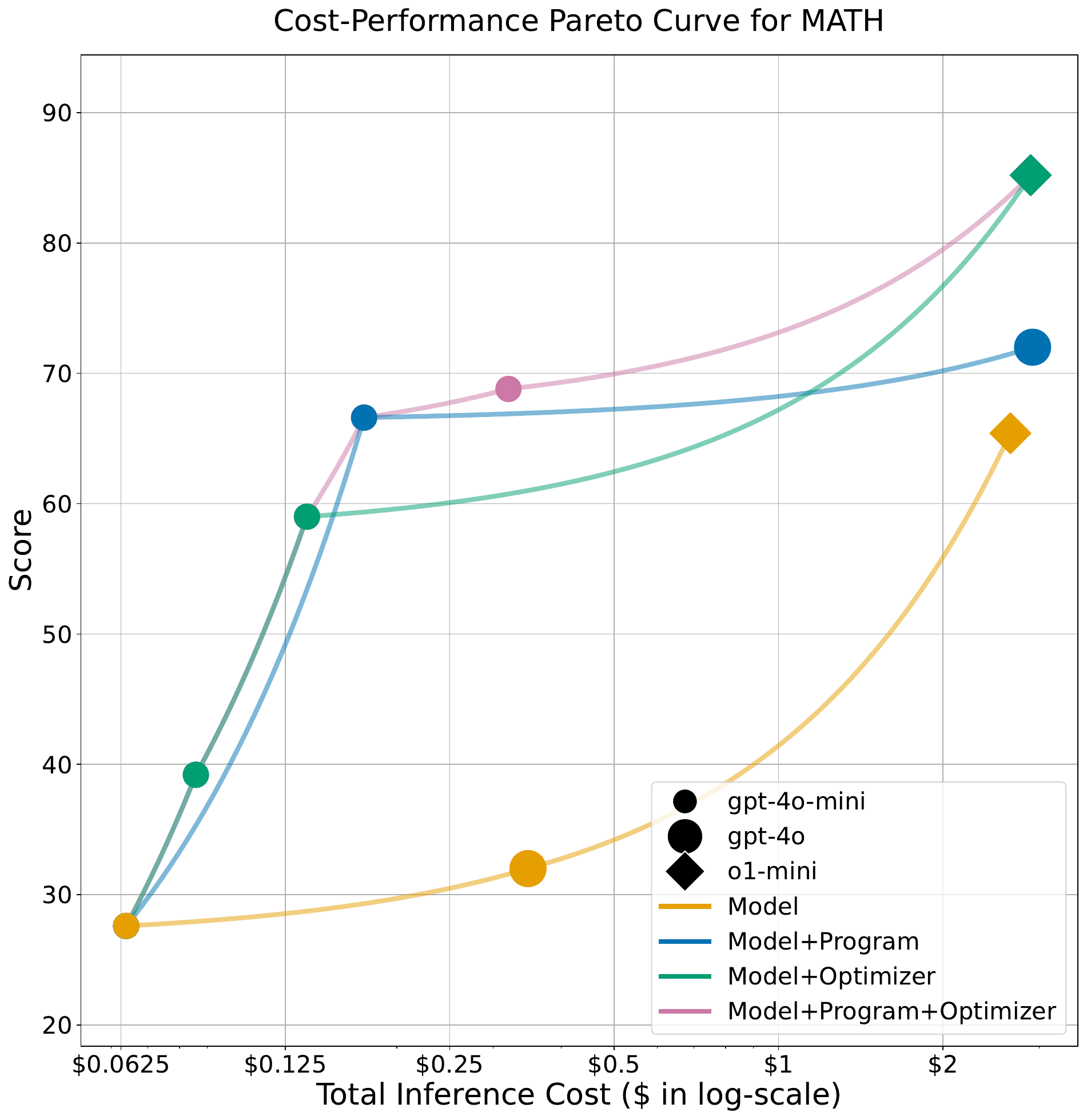}
        \caption{MATH}
        \label{fig:pareto_math}
    \end{subfigure}
    \begin{subfigure}{0.45\linewidth}
        \includegraphics[width=\linewidth]{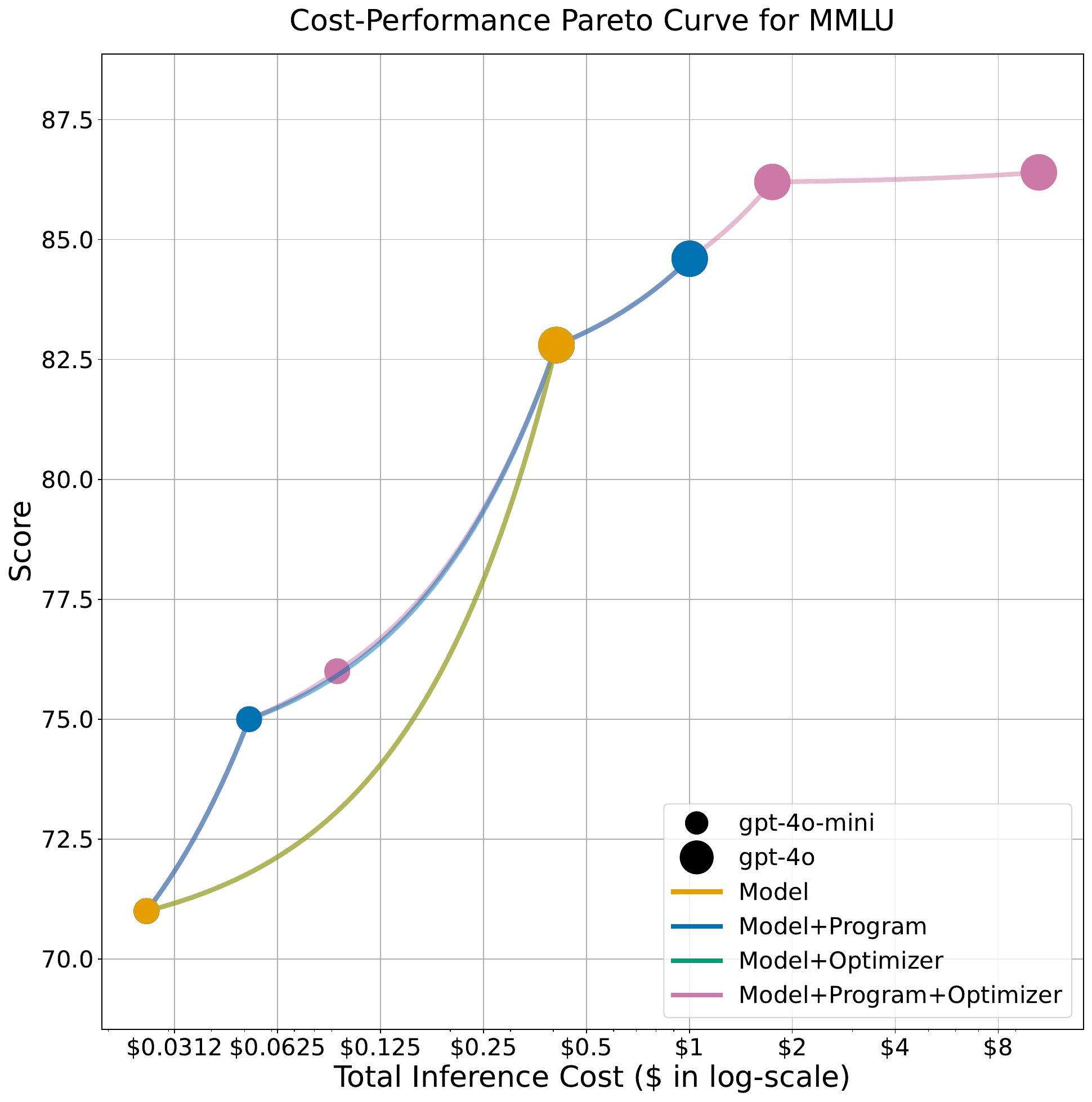}
        \caption{MMLU}
        \label{fig:pareto_mmlu}
    \end{subfigure}\\
    \begin{subfigure}{0.45\linewidth}
        \includegraphics[width=\linewidth]{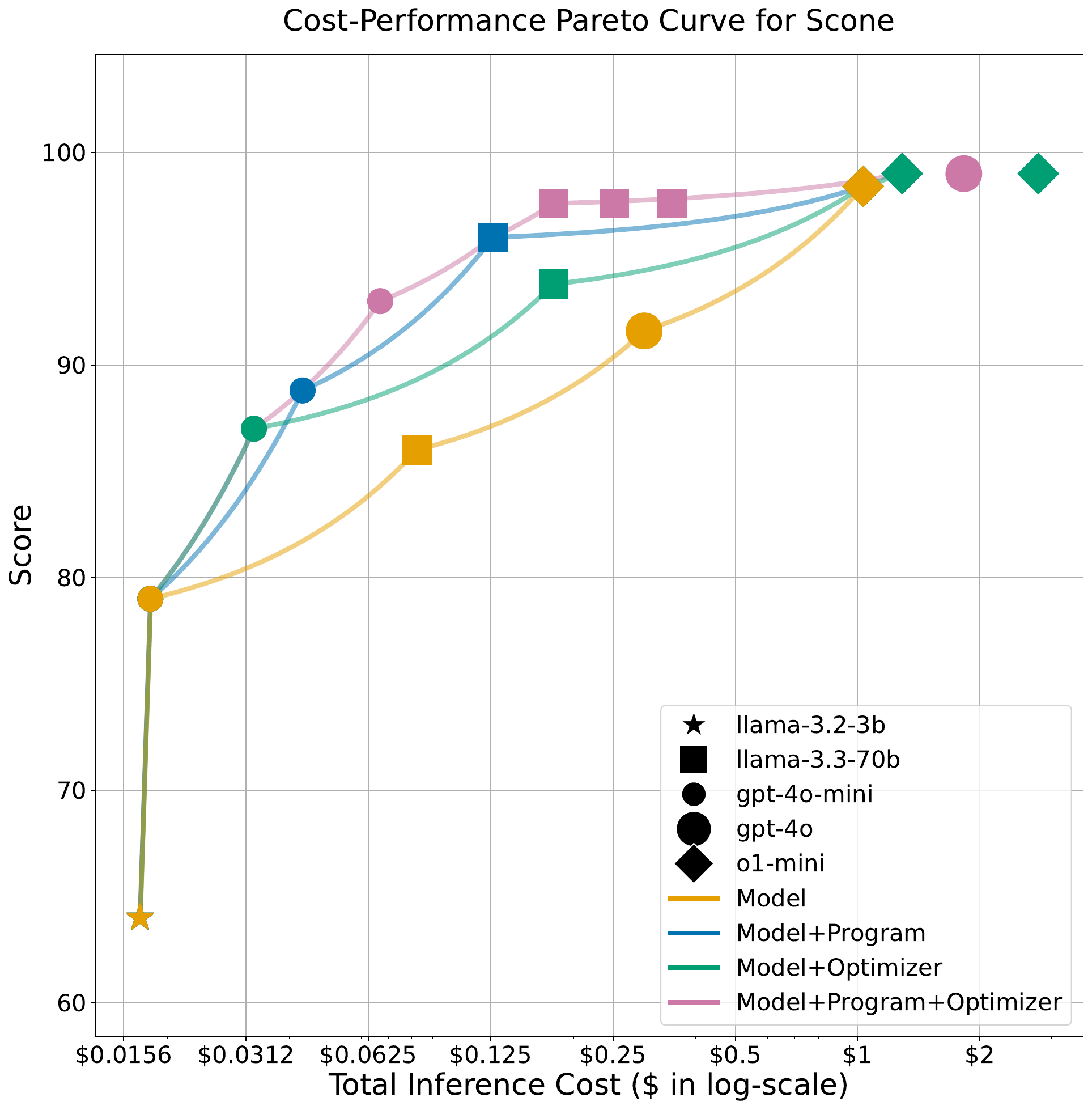}
        \caption{Scone}
        \label{fig:pareto_scone}
    \end{subfigure}
    \begin{subfigure}{0.45\linewidth}
        \includegraphics[width=\linewidth]{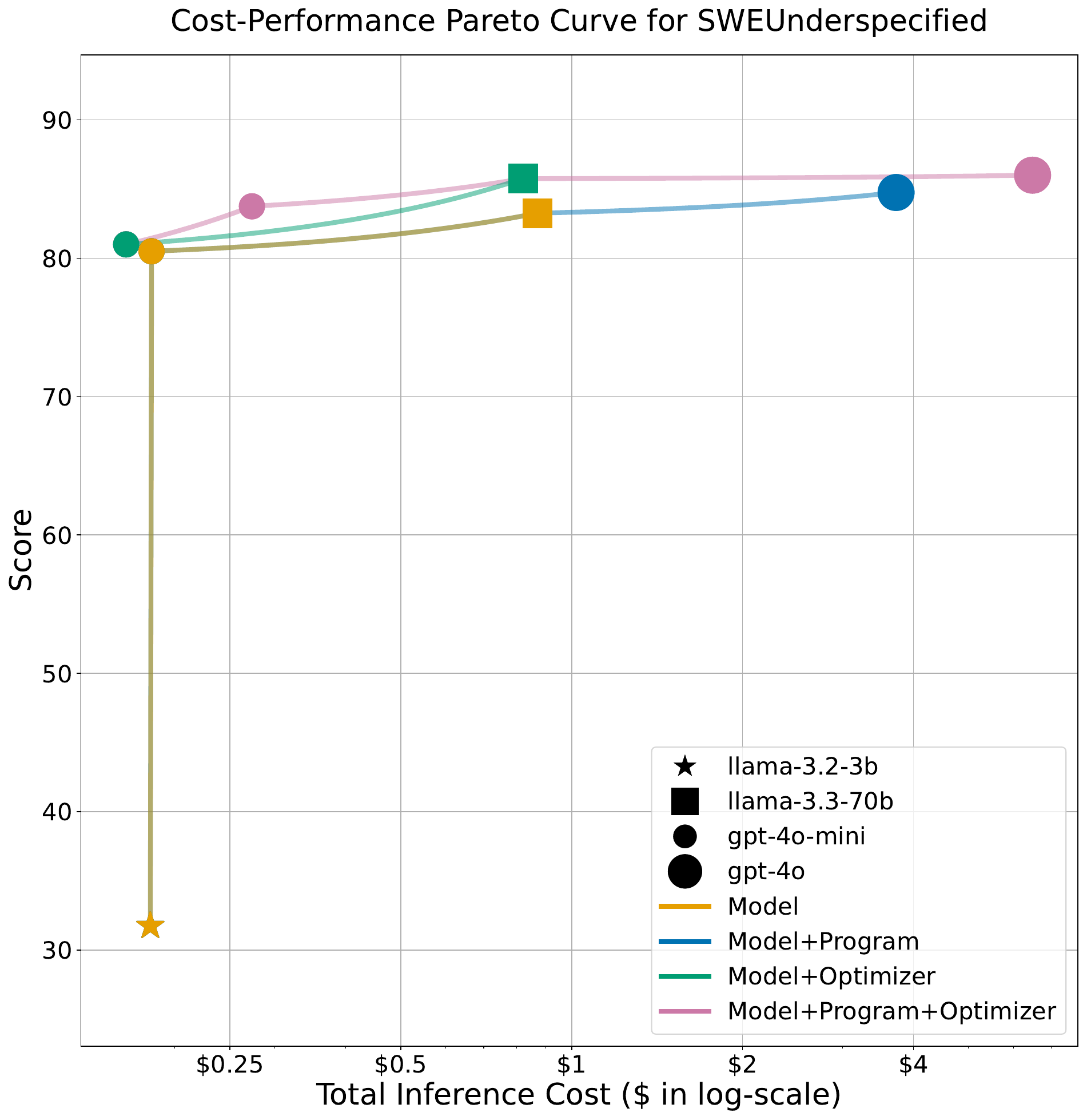}
        \caption{SWEUnderspecified}
        \label{fig:pareto_sweunderspecified}
    \end{subfigure}
    \caption{Performance (Y) vs. Cost (X) Graph for different Benchmarks, Language Programs and Optimizers}\label{fig:pareto_indv2}
\end{figure*}

Figure~\ref{fig:pareto} visualizes the Performance-Cost Pareto optimal configuration aggregated over 7 datasets in LangProBe (Scone, HotpotQA, HumanEval, HeartDisease, Judge, Iris, HotpotQAConditional) and Figures~\ref{fig:pareto_indv3},~\ref{fig:pareto_indv1}, and~\ref{fig:pareto_indv2} and visualize the Performance-Cost Pareto optimal configurations for individuals datasets in LangProBe.

\section{Hyperparameter Settings for Optimizers}

\begin{table*}[ht!]
\small
\centering
\renewcommand{\arraystretch}{1.2}
\definecolor{headerGreen}{RGB}{189,215,138}

\begin{tabular}{|p{2.6cm}|p{8.9cm}|}
\hline
\rowcolor{headerGreen}
\multicolumn{2}{|l|}{\textbf{BootstrapFewShot}}\\
\hline
\textbf{Init Args} &
\texttt{max\_errors=5000 max\_labeled\_demos=2} \\
\hline
\textbf{Compile Args} &
\texttt{(none)} \\
\hline

\rowcolor{headerGreen}
\multicolumn{2}{|l|}{\textbf{BootstrapFewShotWithRandomSearch}}\\
\hline
\textbf{Init Args} &
\texttt{max\_errors=5000 max\_labeled\_demos=2 num\_threads=16} \\
\hline
\textbf{Compile Args} &
\texttt{(none)} \\
\hline

\rowcolor{headerGreen}
\multicolumn{2}{|l|}{\textbf{MIPROv2-lite}}\\
\hline
\textbf{Init Args} &
\texttt{max\_errors=5000 auto="medium" num\_threads=16} \\
\hline
\textbf{Compile Args} &
\begin{tabular}[t]{@{}l@{}} 
\texttt{num\_trials=20} \\
\texttt{max\_bootstrapped\_demos=4} \\
\texttt{max\_labeled\_demos=2}
\end{tabular} \\
\hline

\rowcolor{headerGreen}
\multicolumn{2}{|l|}{\textbf{MIPROv2}}\\
\hline
\textbf{Init Args} &
\texttt{max\_errors=5000 num\_threads=16 num\_candidates=12} \\
\hline
\textbf{Compile Args} &
\begin{tabular}[t]{@{}l@{}} 
\texttt{num\_trials=50} \\
\texttt{max\_bootstrapped\_demos=4} \\
\texttt{max\_labeled\_demos=2} \\
\texttt{batch\_size=35} \\
\texttt{batch\_full\_eval\_steps=5}
\end{tabular} \\
\hline

\rowcolor{headerGreen}
\multicolumn{2}{|l|}{\textbf{RuleInfer-lite}}\\
\hline
\textbf{Init Args} &
\texttt{max\_errors=5000 num\_candidates=10 num\_rules=10 num\_threads=8} \\
\hline
\textbf{Compile Args} &
\texttt{(none)} \\
\hline

\rowcolor{headerGreen}
\multicolumn{2}{|l|}{\textbf{RuleInfer}}\\
\hline
\textbf{Init Args} &
\texttt{max\_errors=5000 num\_candidates=10 num\_rules=20 num\_threads=8} \\
\hline
\textbf{Compile Args} &
\texttt{(none)} \\
\hline
\end{tabular}

\caption{Summary of default optimizer configurations.}
\label{tab:optimizers}
\end{table*}

Table~\ref{tab:optimizers} summarizes the optimizer configurations used in LangProBe.

\section{Introducing RuleInfer Optimizer}

\begin{algorithm}[t]
\caption{\textsc{RuleInfer}: Inducing Rules from Few-Shot Demonstrations to Optimize LM Programs}
\label{alg:ruleinfer}
\begin{algorithmic}[1]
\Require Initial program $\Phi$, Training set $X$, Validation set $X'$, Candidates $N$
\State $\Phi^* \gets \text{ApplyFewShots}(\Phi, X)$ \quad \textcolor{gray}{}
\State $\mu^* \gets$ Evaluate($\Phi^*, X'$) \quad \textcolor{gray}{$\triangleright$ Compute initial score}
\For{$n = 1,\dots,N$}
    \State $\Phi_n \gets$ Apply rule induction to $\Phi^*$ using $X$
    \State $\mu_n \gets$ Evaluate($\Phi_n, X'$)
    \If{$\mu_n > \mu^*$} 
        \State $\Phi^* \gets \Phi_n$ \quad \textcolor{gray}{$\triangleright$ Update best program}
        \State $\mu^* \gets \mu_n$ \quad \textcolor{gray}{$\triangleright$ Update best score}
\EndIf
\EndFor\\
\Return $\Phi^*$ \quad \textcolor{gray}{$\triangleright$ Final optimized program}
\end{algorithmic}
\end{algorithm}

\begin{figure}[th!]
    \centering
    \includegraphics[width=0.95\linewidth]{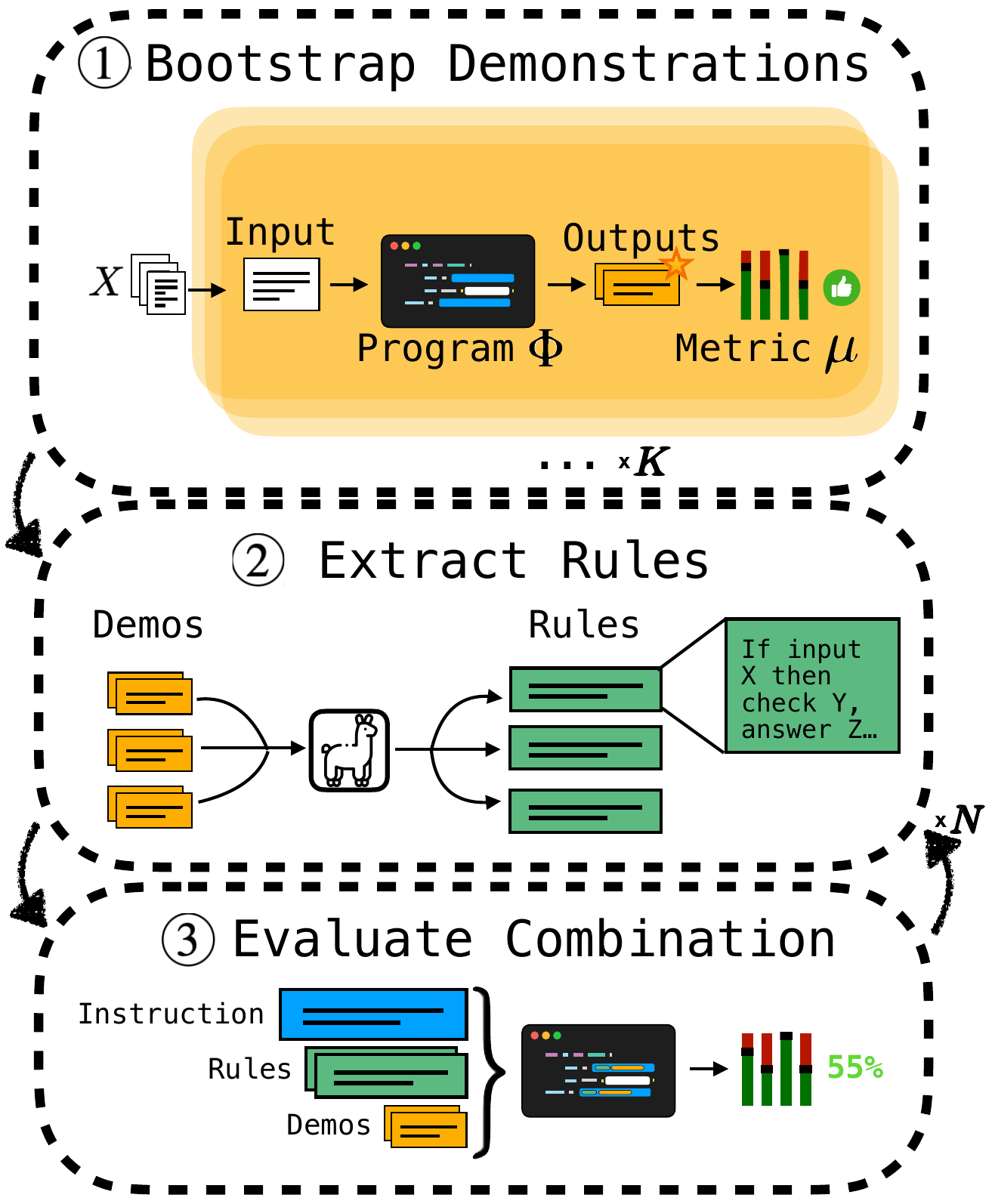}
    \caption{RuleInfer Optimizer Diagram.  RuleInfer Bootstraps demonstrations, extracts rules, and finally finds useful combinations of the rules and demonstrations that work well on the validation set.  Optimizer diagram styles from \cite{opsahlong2024optimizinginstructionsdemonstrationsmultistage}}
    \label{fig:rule-infer}
\end{figure}

We introduce RuleInfer, a prompt optimization approach that identifies actionable rules to optimize the DSPy program performance.  We provide a diagram of RuleInfer in Figure~\ref{fig:rule-infer}. Building upon the BootstrapFewShot optimizer, RuleInfer leverages an LLM to perform rule induction on the generated successful few-shot demonstrations, ascertaining specific insights grounded by the "positive" behavior from the examples to improve model performance. The optimizer then appends these sets of rules to the original task instructions and produces candidates of optimized prompts with natural language guidelines that are validated iteratively with the final output of a best-performing optimized program. 

In comparison to optimizers like BootstrapFewShot (designed mainly for producing examples) and MIPRO (which proposes both instructions and few-shots), RuleInfer offers a distinct way of incorporating grounded, actionable rules derived from existing few-shots. 
RuleInfer excels in tasks with clear, discrete constraints, such as classification (HeartDisease and Iris) or coding domains (HumanEval and SWEBench),  where well-defined rules can be leveraged to create structured decision boundaries for the language model to consider and hence reinforce both consistency and accuracy in performance. This quality makes RuleInfer particularly effective at aligning model behavior to task requirements without extensive manual engineering of the prompt instructions specifically, as the LLM performing rule induction strengthens this alignment. However, the optimizer lacks benefits on tasks are less domain-specific like question-answering tasks (HotPotQA, MMLU) as the induced rules tend can be too broad for open-ended general knowledge queries.

RuleInfer demonstrates how the LangProBe benchmark can be leveraged to introduce and validate new prompt optimization techniques across various tasks. By comparing RuleInfer across existing DSPy optimizers across multiple tasks, programs and models, we provide a streamlined analysis to easily pinpoint instances where rule-based induction excels versus where it underperforms.
This will greatly benefit both prompt optimization developers in benchmarking novel techniques against previous ones and prompt workflow developers in understanding when to use what kinds of optimizers based on their task and program.

\end{document}